\pdfoutput=1
\documentclass[11pt]{article}

\author{
    Tao He$^{1}$\thanks{~~Work was done during an internship at SMU.}, 
    \textbf{
    Lizi Liao$^{2}$, Yixin Cao$^{3}$, Yuanxing Liu$^{1}$, Ming Liu$^{1,4}$\thanks{~~Corresponding Author.},Zerui Chen$^{1}$, Bing Qin$^{1,4}$
    } \\
    $^{1}$Harbin Institute of Technology, Harbin, China \\
    $^{2}$Singapore Management University, Singapore \\
    $^{3}$School of Computer Science, Fudan University  \\
    $^{4}$Peng Cheng Laboratory, Shenzhen, China \\
    \{the, yxliu, mliu, zrchen, qinb\}@ir.hit.edu.cn, \\
    lzliao@smu.edu.sg, caoyixin2011@gmail.com
}

\usepackage{ACL2023}

\usepackage{times}
\usepackage{latexsym}

\usepackage[T1]{fontenc}
\usepackage[utf8]{inputenc}

\usepackage{microtype}

\usepackage{inconsolata}
\usepackage{algorithm}
\usepackage{algorithmic}
\usepackage{amsmath}
\usepackage{amssymb}
\usepackage[inline]{enumitem}
\usepackage{subfigure}
\usepackage{graphicx}
\usepackage{multirow}%
\usepackage{float}
\usepackage{booktabs}
\usepackage{subcaption}
\usepackage{hyperref}

\usepackage{CJKutf8}

\title{Planning Like Human: A Dual-process Framework for Dialogue Planning}

\begin{document}
\maketitle
\begin{abstract}

In proactive dialogue, the challenge lies not just in generating responses but in steering conversations toward predetermined goals, a task where Large Language Models (LLMs) typically struggle due to their reactive nature.
%In proactive dialogue, the core challenge extends beyond mere response generation to actively guiding conversations towards specific objectives—a capability where Large Language Models (LLMs) typically struggle due to their reactive nature.
%Therefore, simply prompting Large Language Models (LLMs) often falls short due to their nature of passively guiding by user instructions.
Traditional approaches to enhance dialogue planning in LLMs, ranging from elaborate prompt engineering to the integration of policy networks, either face efficiency issues or deliver suboptimal performance. 
% Drawing inspiration from the dual-process theory in psychology, which distinguishes between intuitive (fast) and analytical (slow) thinking modes, we introduce the Dual-Process Dialogue Planning (DPDP) framework. DPDP embodies this dual approach through two complementary planning systems: an instinctive policy model for familiar contexts and a deliberative Monte Carlo Tree Search (MCTS) mechanism for complex, novel scenarios.
Inspired by the dual-process theory in psychology, which identifies two distinct modes of thinking—intuitive (fast) and analytical (slow), we propose the Dual-Process Dialogue Planning (DPDP) framework. DPDP embodies this theory through two complementary planning systems: an instinctive policy model for familiar contexts and a deliberative Monte Carlo Tree Search (MCTS) mechanism for complex, novel scenarios.
This dual strategy is further coupled with a novel two-stage training regimen: offline Reinforcement Learning for robust initial policy model formation followed by MCTS-enhanced on-the-fly learning, which ensures a dynamic balance between efficiency and strategic depth. 
Our empirical evaluations across diverse dialogue tasks affirm DPDP's superiority in achieving both high-quality dialogues and operational efficiency, outpacing existing methods.\footnote{~Code available at: \href{https://github.com/cs-holder/DPDP.git}{https://github.com/cs-holder/DPDP.git}}

\end{abstract}

\section{Introduction}
% Large Language Models (LLMs)~\cite{Ouyang2022TrainingLM, Touvron2023LLaMAOA, Touvron2023Llama2O} have proven remarkably adept at comprehending context and producing appropriate responses in a range of dialogue scenarios~\cite{Bang2023AMM, Zhang2023SGPTODBT, Zhao2023IsCE}. However, as LLMs are primarily trained to passively follow users' instructions, LLM-powered dialogue agents often encounter difficulties in proactively guiding the conversation toward a predefined objective~\cite{deng2023plug}. In such scenarios, enhancing the dialogue planning capability of LLM-powered agents becomes pivotal~\cite{zhang2023ask, fu2023improving}.
%Large Language Models (LLMs)~\cite{Ouyang2022TrainingLM, Touvron2023LLaMAOA, Touvron2023Llama2O} have demonstrated remarkable proficiency in understanding context and generating appropriate responses across various dialogue scenarios~\cite{Bang2023AMM, Zhang2023SGPTODBT, Zhao2023IsCE}. However, since LLMs are primarily trained to passively follow users' instructions, dialogue agents powered by LLMs often struggle to proactively steer the conversation toward a predefined objective~\cite{deng2023plug}. Therefore, investigating the dialogue planning task is deemed highly necessary~\cite{Levin1997LearningDS, Liu2018DialogueLW, Cheng2022ImprovingME}.

Large Language Models (LLMs), such as those described by \cite{Ouyang2022TrainingLM,Touvron2023LLaMAOA, Touvron2023Llama2O}, have revolutionized the field of natural language processing by demonstrating an unprecedented ability to understand context and generate coherent responses across a wide range of dialogue scenarios~\cite{Bang2023AMM, Zhang2023SGPTODBT, Zhao2023IsCE}. Despite these advances, LLMs predominantly operate in a reactive mode, often struggling to proactively guide conversations towards specific goals, a critical limitation for achieving truly dynamic interactions~\cite{deng2023plug}. This gap underscores the urgent need for research into dialogue planning mechanisms that can strategically direct conversations, a topic that has been explored but remains an ongoing challenge~\cite{Levin1997LearningDS, Liu2018DialogueLW, Cheng2022ImprovingME}.

\begin{figure}[t]
    \centering
    \includegraphics[width=0.95\linewidth]{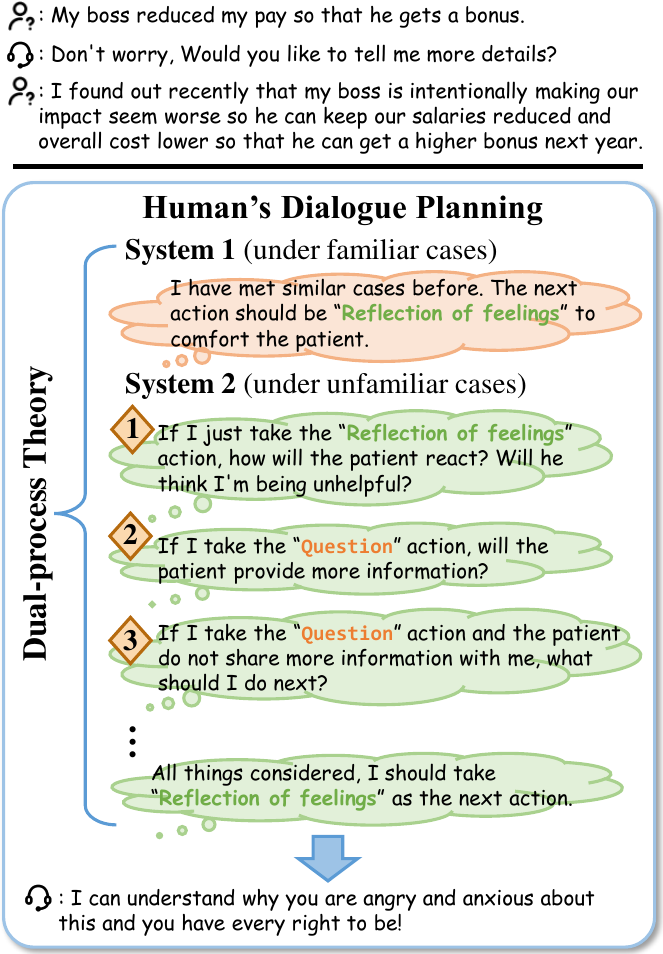}
    \caption{Using dual-process theory for dialogue planning in the human cognitive process. This is a case from ESConv~\cite{Liu2021TowardsES}. ``Question'' and ``Reflection of feelings'' are pre-defined dialogue actions in ESConv.} 
    \vspace{-0.3cm}
    \label{fig:intro_example}
\end{figure}

Dialogue planning, essential for shaping the trajectory of conversations to achieve desired outcomes, has seen various approaches~\cite{Zhang2020LearningGD}. Some studies aim to design more effective prompting procedures, such as Monte Carlo Tree Search (MCTS)~\cite{Vth2023ConversationalTS, Yu2023PromptBasedMT}, resulting in outstanding performance achievements. However, these methodologies frequently encounter inefficiencies deemed unacceptable due to their complex and iterative nature. In contrast, studies exemplified by PPDPP~\cite{deng2023plug} employ Reinforcement Learning (RL) to train a pluggable small model as the policy network, circumventing the high-cost problem of optimizing LLMs and the low-efficiency issue of iterative prompting. However, in practical applications, the trained policy network predicts dialogue actions solely based on the current dialogue history while neglects potential user reactions in subsequent turns, hence resulting in limited performance.

Inspired by the dual-process theory of human cognition, as elaborated by Kahneman~\cite{Kahneman2003MapsOB}, which posits the existence of two distinct modes of thinking—System 1: fast, intuitive, and System 2: slow, analytical—we propose leveraging this framework for dialogue planning. Human conversationalists seamlessly integrate these systems, employing rapid, instinctual responses or engaging in deliberate, strategic thought as situations demand. This theory offers a compelling lens through which to reimagine dialogue planning, suggesting that a blend of intuitive and analytical planning could vastly enhance LLMs' ability to conduct proactive dialogues.

In response, we introduce the Dual-Process Dialogue Planning (DPDP) framework, a novel approach that incorporates two complementary planning systems: a neural policy LM model (System 1) for quick, instinctive responses to familiar situations, and an MCTS-based planner (System 2) for analytic, rational but slow planning in complex or novel scenarios. This framework allows for dynamic switching between systems based on policy LM's uncertainty, %based on the context's uncertainty, 
optimizing for both efficiency and depth of strategy. 
Key to the success of DPDP is the enhancement of the policy model's capability, which we address through a pioneering two-stage training approach. Initially, we employ offline RL to refine the policy model's base, mitigating the impact of suboptimal strategies and noise prevalent in training datasets. Subsequently, we leverage MCTS simulations to guide the policy model towards generating superior strategies, thereby accelerating its convergence and enhancing overall performance. Our comprehensive evaluation across various proactive dialogue tasks unequivocally demonstrates DPDP's superiority over contemporary methodologies, establishing new benchmarks in dialogue planning efficiency and efficacy.

In summary, our contributions are threefold:
\begin{itemize}[leftmargin=*,nosep]
    \item We present a dual-system approach to dialogue planning that mirrors human cognitive processes, balancing efficiency and strategic depth. 
    \item We develop a novel two-stage training method for the policy model, integrating offline RL and MCTS to significantly enhance its performance.
    % We design two-stage training methods for the policy LM that pretrain it using the offline RL paradigm and then further optimize it guided by MCTS.
    \item Experimental results across two datasets validate that our proposed framework effectively outperforms a series of baselines and performs more efficiently than MCTS-based methods.
\end{itemize}

\section{Related Work}
\subsection{LLM-powered Dialogue Policy Planning}
Dialogue planning, critical for guiding systems in task-oriented dialogues to achieve specific goals~\cite{Zhang2020LearningGD}, has been extensively explored~\cite{Jang2020BayesAdaptiveMP, Takanobu2020MultiAgentTD, jang2022gpt, feng2023fantastic, Wang2023DialoguePV}. Despite advancements, the transition to leveraging Large Language Models (LLMs) introduces new challenges, primarily due to their static parameters and constrained capability for long-term planning~\cite{Yao2023RetroformerRL, Xu2023SmallMA}. In response, several methods have been developed, ranging from intricate prompt engineering that encourages self-reflection~\cite{deng2023prompting, zhang2023ask}, to iterative feedback loops for planning enhancement~\cite{fu2023improving}, and even Monte Carlo Tree Search (MCTS) for identifying optimal actions~\cite{Yu2023PromptBasedMT}. While these approaches achieve notable results, they are often marred by inefficiency and high operational costs. An alternative, presented by \citet{deng2023plug}, is the Reinforcement Learning (RL)-based dialogue planning method PPDPP, which utilizes a compact model as a policy network for strategy prediction~\cite{Li2023GuidingLL}, directly using dialogue history. Although this method offers efficiency, it falls short in simulating the nuanced human-like cognitive processes in proactive dialogue, particularly in anticipating the impact of actions on future dialogues. Our proposed solution draws from dual-process theory~\cite{Sun2002DualityOT, Kahneman2003MapsOB}, marrying a policy LM planner for immediate strategy predictions with an MCTS planner for in-depth state simulations. This hybrid approach dynamically alternates between planners based on the system's confidence level, striking a balance between efficiency and strategic effectiveness.

\begin{figure*}[t]
    \centering
    \subfigure[Overall Procedure of DPDP\label{a}]{
        \includegraphics[width=0.51\linewidth]{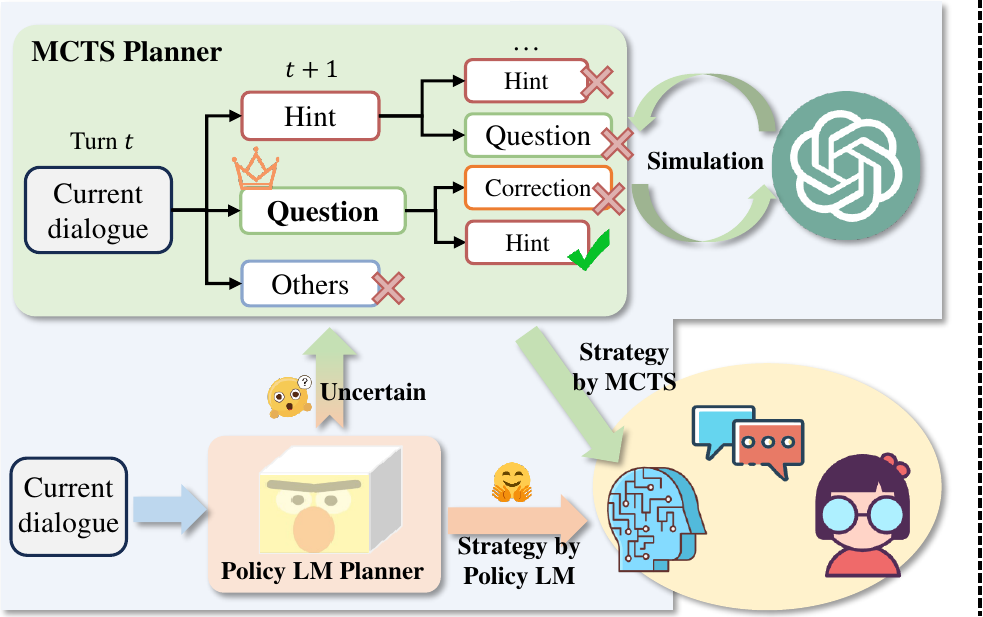}
    }
    %8.7cm
    \subfigure[Two-stage Training for Policy LM \label{b}]{
        \includegraphics[width=0.45\linewidth]{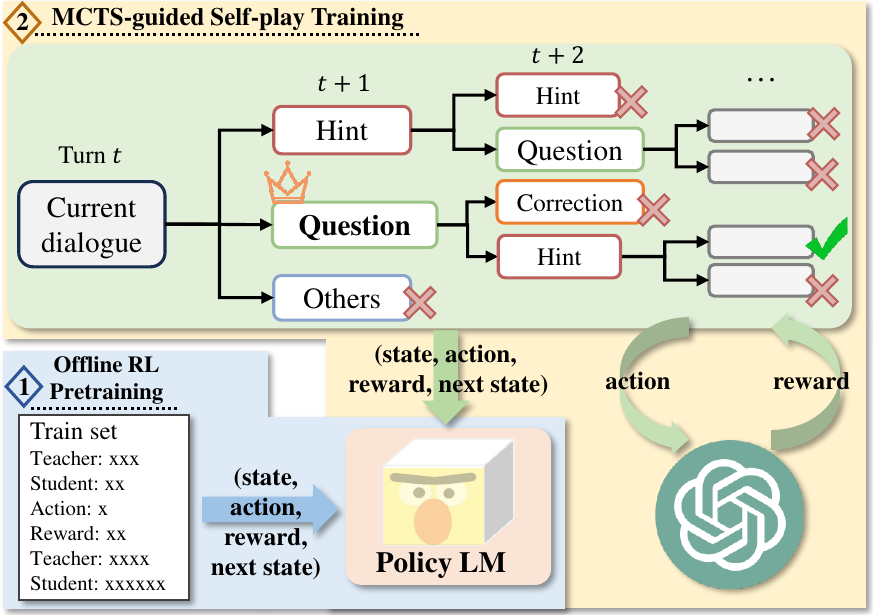}
    }
    %8.3cm
    \caption{DPDP combines two planners: (a) a Policy LM for quick responses and an MCTS planner for complex scenarios, switching when Policy LM is uncertain during inference.
    %DPDP is comprised of two planners: the Policy LM planner and the MCTS planner. (a) For inference, the system operates according to the dual-process theory: the Policy LM planner is prioritized for use for swift responses. However, if the Policy LM planner feels uncertain about the current dialogue state, a switch to the MCTS planner occurs.
    % , which requires deeper contemplation but results in slower responses. 
    (b) We propose a two-stage training approach for Policy LM, involving pretraining over static data with offline RL and further finetuning with MCTS simulations.}
    \label{fig.main}
\vspace{-0.3cm}
    % \vskip-10pt
\end{figure*}

\subsection{Applications of Dual-process Theory}
% Various studies have been actively integrating System 1 and System 2 from dual-process theory~\cite{Kahneman2003MapsOB} into machine learning methodologies. For instance, \cite{Mittal2017ThinkingFA} liken the vector space model and reasoning in knowledge graphs to fast thinking and slow thinking, respectively, proposing a hybrid query processing engine for search applications. \cite{Bengio2017TheCP} introduces a consciousness prior theory to learn high-level concepts, highlighting the close relationship between System 2 abilities and consciousness. In a similar vein, \cite{Chen2019DeepRN} propose an end-to-end framework comprising a generative decoder (fast thinking) and a reasoning module (slow thinking) to tackle complex tasks. Building upon this, \cite{Liu2022ANA} presents a neural network as the fast System 1 and a symbolic module as the slow System 2 to address question-answering tasks. Motivated by dual-process theory, we implement our dialogue planning framework with a fast policy network planner and a slow MCTS planner.
% \yxliu{Add description of dual-process.}
Existing studies have actively integrated System 1 and System 2 from dual-process theory~\cite{Kahneman2003MapsOB} into machine learning methodologies. \citet{Mittal2017ThinkingFA} view the vector space model and reasoning in knowledge graphs as fast and slow systems, respectively.
% , proposing a hybrid query processing engine for search applications. 
% \citet{Bengio2017TheCP} introduces a theory of consciousness prior to learn high-level concepts, 
\citet{Bengio2017TheCP} highlights the close relationship between System 2 abilities and consciousness. 
Additionally, \citet{Chen2019DeepRN} propose an end-to-end framework consisting of a generative decoder (fast thinking) and a reasoning module (slow thinking) to tackle complex tasks. 
Expanding on this, \citet{Liu2022ANA} combine a neural network and a symbolic module by dual-process theory to address question-answering tasks. 
Inspired by above studies, we implement our dialogue planning framework with a fast policy LM planner and a slow MCTS planner.

\subsection{Integrated Learning of RL and MCTS}
In the field of RL, numerous investigations have converged on the amalgamation of MCTS and RL algorithms. Notably, some endeavors aim to leverage RL methodologies to enhance MCTS efficacy~
\cite{Guo2016DeepLF, Anthony2019PolicyGS, Soemers2019LearningPF, Dieb2020OptimizationOD}. For instance, \citet{Guo2016DeepLF} design reward-bonus functions to augment MCTS. \citet{Anthony2019PolicyGS} advocate the utilization of policy networks for the refinement of local policies and investigate planning without an explicit tree search. Concurrently, many studies adopt MCTS as a guiding mechanism for RL training, a paradigm commonly referred to as expert iteration~\cite{Efroni2018HowTC, Anthony2017ThinkingFA, Grill2020MonteCarloTS}. Prominent exemplars of this methodology encompass AlphaZero~\cite{Silver2017MasteringCA, Silver2017MasteringTG} and MuZero~\cite{Schrittwieser2019MasteringAG}. 
Inspired by AlphaZero, we apply the MCTS to enhance the policy LM in the dialogue planning task. 
% Besides, we synergistically leverage both the MCTS and policy LM instead of discarding MCTS like AlphaZero during inference.

\section{Methodology}
\subsection{Preliminaries}\label{sec:preliminaries}
\noindent\textbf{Problem formalization.}
Following existing work~\cite{Wang2020TaskCompletionDP}, we formulate the dialogue process as a Markov Decision Process (MDP). At each turn $t$, according to the observation on the current state $s_t$, i.e. the dialogue history $\{u^{sys}_1, u^{usr}_1,..., u^{sys}_{t}, u^{usr}_{t}\}$, the dialogue system selects an action $a_t\in \mathcal{A}$, where $u^{sys}$ and $u^{usr}$ represent the system and user utterances, $\mathcal{A}$ is a set of candidate strategies pre-defined by domain experts. Then, guided by this action $a_t$, the system player generates the utterance $u^{sys}_{t+1}$. In return, the user player responds to the system with $u^{usr}_{t+1}$. This process repeats until the dialogue goal is achieved or the maximum number of turns $T$ is reached. The objective is to learn a policy $\pi_\theta$ maximizing the expected cumulative rewards over observed dialogue episodes as:
\vspace{-0.1cm}
\begin{equation}
    \pi^*=\arg \max_{\pi_\theta}\bigg[\sum_{t=0}^Tr(s_t,a_t)\bigg],
    \label{eq:total_target}
\end{equation}
where $r(\cdot)$ is a reward function, abbreviated as $r_t$.
\vspace{+0.2cm}

\noindent\textbf{LLM-powered role simulation.}
Following previous work~\cite{deng2023plug}, we employ two LLMs as a user and an assistant to simulate dynamic user-assistant interactions. Role descriptions, along with instructions about their corresponding conversational goals, are delivered to each LLM. Furthermore, we prompt an LLM, following PPDPP~\cite{deng2023plug}, to act as a critic for evaluating the dialogue states. The prompts utilized for role simulation and state evaluation processes remain consistent with those employed in PPDPP. 
Please refer to Appendix~\ref{app:prompting_details} for detailed prompts.
Through this approach, we can concentrate our research efforts on effectively planning strategies for each dialogue turn.

\subsection{Dual-process Planning Framework}
We present the dual-process dialogue planning framework in Figure~\ref{a}. Motivated by human cognitive research, human cognition and behavior are propelled by two cognitive systems: the intuitive and analytic systems. In our study, a smaller model is trained to function as an intuitive policy LM (we implement using RoBERTa-large), capable of directly predicting the next conversational action based on the dialogue history. MCTS is applied as an analytic process by iteratively simulating subsequent dialogue turns to select an approximately optimal strategy through multiple simulations. 
% Priority is given to utilizing the policy LM for action selection. 
If the policy LM is not confident enough in the current state, we shift to employing MCTS for action planning. We propose a nonparameterized control gate mechanism for deciding the switch.

\subsubsection{Policy LM Planner}
% Due to the expensive optimization cost, we propose employing a small model to control the dialogue policy planning. Specifically, we leverage a tunable pre-trained language model, e.g., RoBERTa-large~\cite{Liu2019RoBERTaAR}, as the dialogue policy planner. By feeding into two different MLP layers, the policy LM planner consists of a policy network $\pi_\theta(a|s)$ to predict the next action and a Q-network $Q_\beta(a|s)$ to evaluate each state. We introduce the Q-network to help train the policy network. Please refer to Section~\ref{sec:offline_rl} for more details. For each turn, we predict the next strategy $a_t$ by feeding the policy network with the dialogue state $s_t$, that is $\pi_\theta(a_t|s_t)$.
We propose utilizing a tunable pre-trained language model, e.g., RoBERTa~\cite{Liu2019RoBERTaAR}, as the dialogue policy planner to control the dialogue process.
Different from previous methods~\cite{deng2023plug}, we involve not only a policy network but also a Q-network. Our design rationale is based on two points: (1) The CIMA training set comprises only dialogue snippets rather than complete dialogue histories. It is inadequate to only learn a policy network in our proposed Offline RL-based pretraining method. An action-value function, i.e., Q-Network, is required to aid in training the policy network. (2) We utilize the LLM as the reward function, that is we view the critic LLM as part of the environment. Modeling the environment is a typical method to reduce interaction while maintaining performance~\cite{Luo2022ASO}.
% Thus, learning a Q-network ensures efficient learning and minimizes the number of calls to the LLM during interaction with the environment.

By connecting two different MLP layers, the policy LM planner comprises a policy network $\pi_\theta(a|s)$ for action prediction and a Q-network $Q_\beta(a|s)$ for state evaluation. 
% The Q-network is introduced to facilitate the training of the policy network. Please consult Section~\ref{sec:offline_rl} for further details. 
At each turn, the subsequent strategy $a_t$ is predicted by inputting the dialogue state $s_t$ into the policy network.
% , represented as $\pi_\theta(a_t|s_t)$.

\subsubsection{MCTS Planner}
% Following GDPZero~\cite{Yu2023PromptBasedMT}, we employ MCTS~\cite{weber2010optimization, Liebana2015OpenLS} to simulate deep-level strategic deliberation, a simulation process typically comprising four stages: Selection, Expansion, Evaluation, and Backpropagation. After multiple simulations, the most frequently applied action during these processes is ultimately chosen as the conversational strategy for the next turn of response. Please refer to GDPZero for more details about MCTS on dialogue planning.

% The relationship between the policy LM and MCTS is not mutually independent. In addition to the forthcoming discussion on how MCTS guides policy LM learning, the policy LM also participates in the simulation process of MCTS by producing prior knowledge. To furnish the action prior distribution for each Selection step, GDPZero~\cite{Yu2023PromptBasedMT} calculates action probabilities through repeated prompting of an LLM. In contrast, we employ the policy LM, trained on the specific domain, to produce the prior probability. This approach not only facilitates the application of learned domain-specific knowledge for enhanced initialization but also reduces the frequency of calls to the LLM, thereby improving efficiency and decreasing cost.
Following GDPZero~\cite{Yu2023PromptBasedMT}, we utilize MCTS~\cite{weber2010optimization, Liebana2015OpenLS} for simulating subsequent strategic deliberation, a process typically involving four stages: Selection, Expansion, Evaluation, and Backpropagation. After multiple simulations, the conversational strategy for the next turn of response is determined based on the action most frequently applied during these processes. For further details about MCTS in dialogue planning, please refer to Appendix~\ref{app:mcts_details}.

% The relationship between policy LM and MCTS is interdependent. Besides the forthcoming discussion on how MCTS guides policy LM learning, policy LM also contributes to the simulation process of MCTS by producing prior knowledge. 
Different from GDPZero, we apply policy LM to produce prior knowledge for MCTS.
To establish the prior action distribution for each Selection step, GDPZero computes action probabilities by repeatedly prompting an LLM. Conversely, we utilize the domain-specific trained policy LM to generate the prior probability. This approach not only enables the application of learned domain-specific knowledge for improved initialization but also diminishes the frequency of calls to the LLM, thus enhancing efficiency and reducing costs.
We verify this operation in the following experiments.

\subsubsection{Synergizing Two Planners}
Our objective is to synergistically utilize both the policy LM planner and MCTS planner to establish an adaptive dual-process system. Similar to human cognitive processes during proactive dialogue, individuals can rapidly respond to appropriate strategies when encountering familiar conversational states. Conversely, when encountering unfamiliar states, it becomes necessary to select the most suitable strategy by simulating potential reactions in subsequent dialogue turns. In our framework, we give priority to utilizing a policy LM for action selection. If the policy LM detects inadequate confidence regarding the current state, we switch to employing MCTS for action planning. 

To reduce reliance on training, we propose a non-parameterized control gate mechanism to control switching.
For the action distribution $\pi_\theta(a_t|s_t)$ predicted by policy LM, we assess the uncertainty by the probability difference for the top-2 values: $\delta(\pi_\theta(a_t|s_t))=top(1) - top(2)$, where $top(i)$ means the $i$-th largest value in $\pi_\theta(a_t|s_t)$.
If this difference surpasses a threshold $\eta$, it indicates that the policy LM has high confidence in the current decision-making, prompting the utilization of the policy LM for action selection. Conversely, small differences imply that the policy LM is uncertain, prompting the utilization of MCTS for action selection. By setting appropriate $\eta$, we can roughly control the ratio of MCTS used. For details on how to determine $\eta$, please see Appendix~\ref{app:determine_mcts_ratio}.
Of course, other uncertainty measures like entropy can also be applied in our framework.
% Additionally, we propose an automatic method to control $\eta$ based on the employed ratio of the MCTS planner. Specifically, we maintain a list to store all previous $\delta(\pi_\theta(a_t|s_t))$ values and compute the next threshold $\eta$ as the quantile of the value list.

% \input{Sections/algorithm_1}

\subsection{Two-stage Training for Policy LM}
\subsubsection{Offline RL-based Pretraining}\label{sec:offline_rl}
% Before performing interactive online learning, an appropriate initialization is expected to reduce the time of interaction.
% Nevertheless, both annotated strategies and dialogue responses can be sub-optimal or noisy, thus direct supervised learning might lead to a biased initialization state. Unlike direct supervised learning on the training set, as in PPDPP~\cite{deng2023plug}, we use offline RL~\cite{Kumar2020ConservativeQF, Kostrikov2021OfflineRL} for initial learning. The advantages of utilizing RL as opposed to direct supervised learning lie in the ability to selectively learn from a dataset by employing soft rewards rather than hard labels, thereby avoiding the learning of suboptimal strategies within the dataset. 
% Specifically, our approach involves initially assigning scores to each dialogue turn in the training set using ChatGPT, serving as rewards for the given dialogue strategies.
% By means of that, we re-construct a MDP corpus containing complete states, actions, and rewards. By employing this corpus, we train the policy LM comprised of the policy network $\pi_\theta(a|s)$ and the Q-network $Q_{\beta}(s,a)$. 
Our two-stage training scheme for the Policy LM, depicted in Figure \ref{b}, begins with offline RL-based Pretraining to ensure effective initialization, aiming to reduce interaction time during subsequent online learning. Unlike direct supervised learning, which can introduce biases through suboptimal or noisy data, offline RL~\cite{Kumar2020ConservativeQF} refines pretraining by using soft rewards to discern valuable strategies from the dataset, avoiding the pitfalls of hard labels. In detail, our approach initially assigns scores to each dialogue turn in the training set using the critic LLM, which serves as rewards for annotated dialogue strategies. As a result, we reconstruct an MDP corpus comprising complete states, actions, and rewards. Utilizing this corpus, we pretrain the policy LM, which consists of the policy network $\pi_\theta(a|s)$ and the Q-network $Q_{\beta}(s,a)$.

Specifically, the pretraining details for ESConv~\cite{Liu2021TowardsES} and CIMA~\cite{Stasaski2020CIMAAL} datasets differ due to the absence of complete dialogue trajectories on CIMA.
We outline the optimization process for ESConv here, while details for CIMA are provided in Appendix~\ref{app:cima_pretraining}.
For the policy network, we optimize it by:
\begin{equation}
    \mathcal{L}_{pre,\theta}=-\sum_{t=1}^{T}\hat{Q}(s_{t},a_{t})\log \pi_{\theta}(a_{t}|s_{t}),
\end{equation}
where $\hat{Q}(s_t,a_t)=\sum_{t=1}^T\gamma^tR(a_t|s_t)$ represents cumulative rewards, $\gamma$ means the discount factor, and $R(a_t|s_t)$ is the received reward by selecting action $a_t$ upon the state $s_t$.
% which is calculated by LLM. 
We use the same strategy as PPDPP, which generates 10 evaluations on the current state, maps each evaluation into a pre-defined score, and finally computes the mean value as the reward~\cite{deng2023plug}. For ESConv, we map ``feel worse'', ``feel the same'', ``feel better'', ``solved'' into -1.0, -0.5, 0.1, 1.0.
For the Q-network $Q_\beta(s,a)$, we optimize it to approximate $\hat{Q}(s,a)$:
% \begin{equation}
\begin{align}
    \mathcal{L}_{pre,\beta}=\sum_{t=1}^T\text{MSE}(Q_\beta(s_t,a_t),\hat{Q}(s_t,a_t)).
\end{align}
% \end{equation}
Finally, the overall optimizing loss for the pretraining stage is:
\begin{equation}
    \mathcal{L}_{pre} = \mathcal{L}_{pre,\theta} + \lambda_1 * \mathcal{L}_{pre,\beta},
\end{equation}
where $\lambda_1$ is a hyperparameter to control loss weight.
By doing this, we expect to learn a better initialization compared to direct supervised learning.

\subsubsection{MCTS-guided Self-play Training} \label{sec:self-play_train}
Additional interaction with the environment is necessary since static training sets cannot cover the entire state-action space.
In interactive online learning, we initiate two LLMs to simulate self-play dialogues between the user and the assistant.
% , mimicking dynamic user-assistant interactions. 
Given the current state $s_t$, rather than directly utilizing the policy agent to predict the next action, we utilize MCTS for action prediction. The predicted action is then mapped to a pre-defined natural language instruction, $\mathcal{M}_a(a_t)$. Subsequently, the dialogue history $s_t$ along with $\mathcal{M}_a(a_t)$ will trigger the LLM to generate the appropriate system response, then prompt the LLM to generate the corresponding user response. Following this, the dialogue process transitions to a new state $s_{t+1}$. We employ an LLM as a critic to calculate the action reward $r_t$. The collected transition records $\{s_t,a_t,s_{t+1},r_t\}$ are used to train the policy model.

We optimize the policy LM using the Actor-Critic algorithm instead of the REINFORCE algorithm~\cite{Sutton1999PolicyGM}. The optimization loss for the Q-network is as follows:
\begin{equation*}
    \begin{aligned}
    \mathcal{L}_{sp,\beta} &= \sum_{t=1}^T[Q^*(s_t, a_t) - Q_\beta(s_{t}, a_{t})]^2,\\
    Q^*(s_t, a_t) &= R(a_t|s_t) + \gamma * \max_{a'}Q_\beta(s_{t+1}, a'),
    \end{aligned}
\end{equation*}
and for the policy network as:
\begin{equation*}
    \begin{aligned}
    \mathcal{L}_{sp,\theta} = \sum_{t=1}^T[&(Q_\beta(s_{t}, a_{t})-\hat{Q}(s_{t}, a_t))\\
    &*\log\pi_\theta(a_t|s_t)].
    \end{aligned}
\end{equation*}
where $Q_\beta$ is the Q-network to calculate state-action values. $\hat{Q}(s_{t}, a_t)$ is cumulative rewards defined before.
Finally, the overall optimizing loss for the self-play training phase is:
\begin{equation}
\mathcal{L}_{sp}=\mathcal{L}_{sp,\theta}+\lambda_2 * \mathcal{L}_{sp,\beta},
\end{equation}
where $\lambda_2$ is also a loss weight.

\section{Experiments}
\subsection{Datasets}
We evaluate the proposed framework on three proactive dialogue datasets, including ESConv~\cite{Liu2021TowardsES} (emotional support dialogue), CIMA~\cite{Stasaski2020CIMAAL} (tutoring dialogue), and CraigslistBargain (or CB, negotiating prices)~\cite{He2018DecouplingSA}.
ESConv is split into 1040\/130\/130 cases for training\/valid\/test set, with pre-defined 8 actions. CIMA is split into 909/113/113 cases for training\/valid\/test set, with 5 dialogue actions. ESConv and CIMA are collaborative dialogue tasks, where both participants share the same goal. In contrast, CB is a non-collaborative dialogue task where the buyer aims for the lowest price, and the seller aims for the highest. CB consists of 3290 training cases, 188 valid cases, and 188 testing cases, involving 11 buyer bargaining actions.
Please refer to Appendix~\ref{app:strategy_prompt} for pre-defined dialogue actions.
% The statistics of adopted datasets are presented in Table~\ref{tab:dataset_statistics}. 
Following the settings of PPDPP~\cite{deng2023plug}, we use human-annotated dialogues in the train set for pretraining. For the self-play training phase, we only use the case background information in the dataset for state initialization.

\subsection{Baselines}
% Our framework is capable of automatically switching between the policy LM and MCTS planner during inference.
% We aim to demonstrate the efficacy of this framework, primarily manifested in two aspects. Firstly, we aim to showcase satisfactory performance solely by utilizing the policy LM without employing MCTS. Therefore, 
We aim to demonstrate the efficacy of this framework by primarily comparing against PPDPP. Additionally, we follow PPDPP to compare with a general fine-tuning dialogue model DialoGPT~\cite{zhang2019dialogpt}, and a range of prompting-based methods, including Standard Prompting, Proactive~\cite{deng2023prompting}, ProCoT~\cite{deng2023prompting}, Ask-an-Expert~\cite{zhang2023ask}, and ICL-AIF~\cite{fu2023improving}. 
Following \citet{deng2023plug}, we report the results of baselines.
% \yxliu{Secondly, by integrating MCTS, we aspire to enable the system to emulate human planning behavior, thereby achieving a trade-off between high efficiency and excellent performance.}

\subsection{Evaluation Metrics}
% \yxliu{We evaluate xxxx with xxxx.}
% For automatic evaluation, we adhere to the employment of two key evaluation metrics, namely the average turn (AT) and the success rate (SR). AT measures the efficiency of goal completion by calculating the average number of turns needed to achieve the goal. SR measures the effectiveness of goal completion by computing the success rate of achieving the goal within a pre-defined maximum turn.
% Besides, upon analyzing specific examples, we discovered biases in directly utilizing ChatGPT for evaluation in ESConv. Therefore, we also conduct human evaluation for accurate assessments. Three annotators are asked to compare the generated responses from DPDP (policy LM) with those from PPDPP. We assess the responses from four primary perspectives, which are \textbf{Sug}gestion, \textbf{Ide}ntification, \textbf{Com}forting, and \textbf{Ove}rall: (1) Suggestion: Which assistant provides more helpful suggestions for solving the problem? (2) Comforting: Which assistant is more skilled at comforting you? (3) Identification: Which assistant is more helpful in exploring and identifying the problem? (4) Overall: Which assistant can better solve the patient's problem? Each annotator is asked to answer whether DPDP (policy LM) wins PPDPP, and the answers include win, lose, and tie on these four aspects. Finally, we average results from three annotators.
For automatic evaluation, we employ two key metrics: the average turn (AT) and the success rate (SR). AT measures goal completion efficiency by calculating the average number of turns required to achieve the goal, while SR measures goal completion effectiveness by computing the success rate of achieving the goal within a predefined maximum number of turns. For CB, following PPDPP, we also use SL (Sale-to-List Ratio) to evaluate the buyer's deal. A higher SL represents the buyer gets more benefits from deals. If the deal fails, we assign SL as 0. Additionally, upon analyzing specific examples, we discovered biases in directly using ChatGPT for evaluation in ESConv. Therefore, we also conduct human evaluation for comparison. 
Three annotators compare the generated responses from four perspectives: Suggestion \textbf{Sug.}, Identification \textbf{Ide.}, Comforting \textbf{Com.}, and Overall \textbf{Ove.}. The instruction for each perspective is presented in Appendix~\ref{app:human_eval_instruction}.
Each annotator is asked to determine whether DPDP (policy LM) outperforms PPDPP, with possible answers being win, lose, or tie for these four aspects. Finally, we average the results from three annotators.

\subsection{Experimental Details}
In line with PPDPP~\cite{deng2023plug}, we instantiate the policy LM utilizing RoBERTa-large~\cite{Liu2019RoBERTaAR}. We implement MCTS using codes from GDPZero~\cite{Yu2023PromptBasedMT}. We also employ gpt-3.5-turbo-0613 as the static LLM for both the playing system and the user, alongside the reward model. All used prompts and temperatures are consistent with PPDPP.
% Regarding the role-playing LLMs, we set the temperature $\tau=0$ to generate deterministic outputs consistent with identical inputs. Meanwhile, for the reward model, we establish $\tau = 1.1$ and iterate 10 times to integrate scalar rewards. The role-playing prompts for the assistant, user players, and reward model remain consistent with those of PPDPP. 
Additionally, we adhere to the same mapping $\mathcal{M}_a$ of natural language instructions with PPDPP for dialogue actions.
For more training details, please refer to the Appendix~\ref{app:implementation_details}.
% For each dialogue, we set the maximize turn as 8. 
% We set $\lambda_1=\lambda_2=1.0$ for ESConv and $\lambda_1=10.0$ and $\lambda_2=1.0$ for CIMA. We pretrain for 10 epochs on ESConv and CIMA with learning rate (lr) of 6e-6, while only self-play train 5 epochs and 3 epochs on ESConv and CIMA with lr=1e-6, respectively. For each self-play epoch, we sample 100 cases from the train set.

% \input{Tables/datasets}
\subsection{Results and Analysis}
\subsubsection{Overall Performance}
\noindent\textbf{Automatic evaluation results.}
% \subsubsection{Automatic Evaluation}
Table~\ref{tab:static_results} \&~\ref{tab:cb_results} summarizes the experimental results of method comparisons on three datasets. On ESConv and CIMA, the proposed DPDP method, integrating both System 1 and System 2, consistently outperforms all the baselines by a noticeable margin. 
% These findings underscore DPDP's capacity not only to effectively accomplish more scenarios (higher Success Rate) but also to adeptly achieve dialogue goals with less Average Turn.
We further analyze the influence of different training methods and systems. Initially, we demonstrate the effectiveness of using System 1 (or Policy LM) alone. Compared to PPDPP, System 1 also utilizes only the policy network to predict subsequent actions. System 1, trained only through pretraining or self-play training, consistently outperformed PPDPP, affirming the efficacy of both offline RL-based pretraining and MCTS-guided self-play training. Furthermore, combining both training methods in System 1 yields further improvements, validating the rationality of the two-stage training approach. System 2, with multiple simulations, achieved superior results compared to System 1, underscoring the effectiveness of MCTS. However, results on CIMA demonstrate that relying solely on MCTS may not lead to optimal outcomes. By appropriately combining System 1 and System 2, better results can be achieved while maintaining higher efficiency (requiring fewer dialogue turns to complete). 

For CraigslistBargain, firstly, compared to the previous state-of-the-art method, DPDP based on System 1 (Policy LM) significantly improves AT (5.57$\rightarrow$ 5.03), SR (0.6649$\rightarrow$ 0.7447), and SL (0.3376$\rightarrow$ 0.4108), demonstrating that our two-stage training method not only enhances deal success rates but also increases benefits. Secondly, System 2 (MCTS) significantly increases the success rates but reduces benefits. This suggests that the System 2 is more prone to compromise. The reason may be that System 2 doesn't require training and terminates the dialogue once the benefits exceed the pre-defined threshold; while the optimized Policy LM focuses on higher-benefits dialogues from MCTS, thereby improving benefits compared with previous work while maintaining a high deal success rate. 
These results also indicate that DPDP needs further improvement for non-collaborative proactive dialogue tasks in the future.

% On the one hand, we observe that utilizing only offline RL pretraining yields excellent performance, surpassing the previous state-of-the-art method.
% % This validates the effectiveness of our initialization approach.
% On the other hand, without pre-training, our direct adoption of MCTS guidance during training already outperformed PPDPP.
% % , indicating the efficacy of MCTS-guided training.
% Subsequent to the two-stage training, further enhancements were observed. 
% % This further validates the rationality of our designed two-stage training approach. 
% % Additionally, alongside the increase in success rate, a significant reduction in Average Turn (AT) indicates improved strategy quality, effectively reducing the application cost of LLMs.
% Additionally, the effectiveness of System 2 is significantly better than that of System 1. However, the results on CIMA also demonstrate that an appropriate combination of System 1 and System 2 can yield superior performances.
% % while requiring fewer LLM invocation instances.

% \yxliu{Although DPDP exhibits impressive performance, upon analyzing specific examples, we discovered biases in directly utilizing ChatGPT for evaluation. Therefore, we also conduct human evaluation for accurate assessments.}

\begin{table}[t]
    \centering
    \renewcommand*{\arraystretch}{1.1}
    \resizebox{\linewidth}{!}{
    \begin{tabular}{lcccc}
    \toprule
    \multirow{2}{*}{\textbf{Models}} & \multicolumn{2}{c}{\textbf{ESConv}} & \multicolumn{2}{c}{\textbf{CIMA}}\\
    & AT$\downarrow$ & SR$\uparrow$ & AT$\downarrow$ & SR$\uparrow$\\
    \midrule
    \multicolumn{1}{l}{DialoGPT~\cite{zhang2019dialogpt}} & 5.31 & 0.7538 & 5.43 & 0.4956 \\
    \midrule
    \multicolumn{1}{l}{Standard} & 5.10 & 0.7692 & 3.89 & 0.6903 \\
    \multicolumn{1}{l}{AnE~\cite{zhang2023ask}} & 4.76 & 0.8000 & 3.86 & 0.6549 \\
    \multicolumn{1}{l}{Proactive~\cite{deng2023prompting}} & 5.08 & 0.7538 & 4.84 & 0.5310 \\
    \multicolumn{1}{l}{ProCoT~\cite{deng2023prompting}} & 4.75 & 0.7923 & 4.58 & 0.5487 \\
    \multicolumn{1}{l}{ICL-AIF~\cite{fu2023improving}} & 4.69 & 0.8079 & 4.19 & 0.6106 \\
    \multicolumn{1}{l}{PPDPP~\cite{deng2023plug}} & \underline{4.56} & \underline{0.8462} & \underline{3.03} & \underline{0.8407} \\
    \midrule
    \midrule
    \multicolumn{1}{l}{DPDP (System 1)} & {3.61} & {0.9000} & \textbf{2.24} & {0.9469} \\
    \multicolumn{1}{l}{$\quad$-w/o PT} & 4.22 & 0.8769 & 2.36 & 0.9292 \\
    \multicolumn{1}{l}{$\quad$-w/o SPT} & 3.97 & 0.8692 & 2.51 & 0.8938 \\
    \multicolumn{1}{l}{{DPDP (System 2)}} & {2.13} & {0.9923} & {2.49} & {0.9735} \\
    \multicolumn{1}{l}{{DPDP (System 1\&2)}} & \textbf{2.13} & \textbf{0.9923} & {2.28} & \textbf{0.9823} \\
    \bottomrule
    \end{tabular}
    }
    \caption{Experimental results on ESConv and CIMA. PT means Pretraining and SPT means Self-Play Training.}
    \label{tab:static_results}
\end{table}

\begin{table}[t]
    \centering
    \renewcommand*{\arraystretch}{1.1}
    \resizebox{\linewidth}{!}{
    \begin{tabular}{lccc}
    \toprule
    \multirow{2}{*}{\textbf{Models}} & \multicolumn{3}{c}{\textbf{CraisglistBargain}}\\
    & AT$\downarrow$ & SR$\uparrow$ & SL$\uparrow$\\
    \midrule
    \multicolumn{1}{l}{DialoGPT~\cite{zhang2019dialogpt}} & 6.73 & 0.3245 & 0.2012 \\
    \midrule
    \multicolumn{1}{l}{Standard} & 6.47 & 0.3830 & 0.1588 \\
    \multicolumn{1}{l}{AnE~\cite{zhang2023ask}} & 5.91 & 0.4521 & 0.2608 \\
    \multicolumn{1}{l}{Proactive~\cite{deng2023prompting}} & 5.80 & 0.5638 & 0.2489 \\
    \multicolumn{1}{l}{ProCoT~\cite{deng2023prompting}} & 6.22 & 0.5319 & 0.2486 \\
    \multicolumn{1}{l}{ICL-AIF~\cite{fu2023improving}} & 6.53 & 0.3617 & 0.1881 \\
    \multicolumn{1}{l}{PPDPP~\cite{deng2023plug}} & 5.62 & 0.6117 & \underline{0.3376} \\
    \multicolumn{1}{l}{$\quad$-w/o SFT} & 5.71 & 0.6223 & 0.3354 \\
    \multicolumn{1}{l}{$\quad$-w/o RL} & \underline{5.57} & \underline{0.6649} & 0.2280 \\
    \midrule
    \midrule
    \multicolumn{1}{l}{DPDP (System 1)} & {5.03} & {0.7447} & \textbf{0.4108}  \\
    \multicolumn{1}{l}{{DPDP (System 1\&2, 22.3\% MCTS)}} & {3.69} & {0.8298} & {0.3102} \\
    \multicolumn{1}{l}{{DPDP (System 1\&2, 51.4\% MCTS)}} & {2.77} & {0.9468} & {0.3118} \\
    \multicolumn{1}{l}{{DPDP (System 1\&2, 60.3\% MCTS)}} & \textbf{2.49} & {0.9681} & {0.2856} \\
    \multicolumn{1}{l}{{DPDP (System 2)}} & 2.78 & \textbf{0.9734} & 0.2728 \\
    \bottomrule
    \end{tabular}
    }
    \caption{Experimental results on CraisglistBargain.}
    \label{tab:cb_results}
\end{table}

% \input{Tables/human_eval}

% \subsubsection{Human Evaluation}
% Following previous studies~\cite{Liu2021TowardsES, Joshi2021DialoGraphII}, we conduct human evaluation on 50 randomly sampled dialogues from ESConv. Three annotators are asked to pair the generated responses by DPDP (policy LM) with PPDPP. We measure four main perspectives of the responses, including \textbf{Sug}gestion, \textbf{Ide}ntification, \textbf{Com}forting, and \textbf{Ove}rall. The instructions for four perspectives are as follows:
\noindent\textbf{Human evaluation results.}
Building upon prior research~\cite{Liu2021TowardsES, Joshi2021DialoGraphII}, we conduct human evaluations on 50 randomly selected dialogues from ESConv.
We focus on ESConv due to its pronounced automatic evaluation bias, primarily stemming from the subjective nature of assessing a patient's state. 
% In the initial human evaluation phases, we also observed individual disparities in evaluating treatment effectiveness and then we standardized criteria accordingly. 
% Conversely, CIMA evaluation focuses solely on verifying the completeness and accuracy of students' translations, posing fewer challenges for LLMs.
We present the results in Figure~\ref{fig:human_eval_esc}. We also provide a human evaluation of CIMA in Appendix~\ref{app:more_human_eval}.
It is evident that our approach differs significantly from PPDPP in the first two criteria: DPDP (System 1) tends to provide advice, whereas PPDPP leans towards expressing empathy, with comparable proficiency in understanding the patient's condition. Given that offering practical advice might be more beneficial than merely expressing empathy in actual psychological counseling, this disparity elucidates why DPDP (System 1) scores considerably higher than PPDPP in the Overall aspect.

\begin{figure}[!t]
    \centering
    \includegraphics[width=1.0\linewidth]{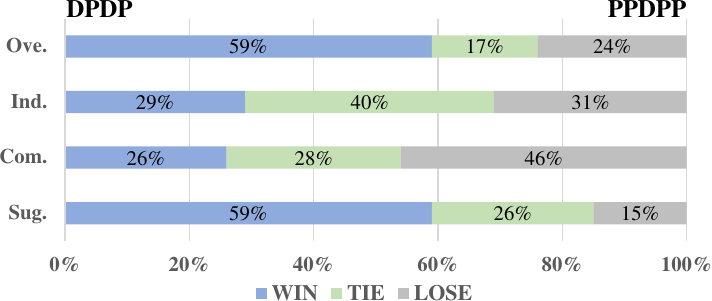}
    \caption{Human evaluation results on ESConv.} 
    \vspace{-0.2cm}
    \label{fig:human_eval_esc}
\end{figure}

% The results indicate that DPDP (policy LM) tends to provide suggestions to solve issues rather than excelling in expressing empathy towards patients. It is precisely due to its ability to offer valuable advice that DPDP (policy LM) also demonstrates better overall curative effects. This is because the primary responsibility of a therapist is to provide solutions for problems rather than solely expressing empathy.
% Each annotator is asked to determine whether DPDP (policy LM) outperforms PPDPP, with response options including win, lose, and tie. The results are presented in Table~\ref{tab:human_eval}. The results indicate that DPDP (policy LM) tends to provide suggestions rather than excel in expressing empathy towards patients. It is precisely due to its ability to offer valuable advice that DPDP (policy LM) also demonstrates better overall curative effects. This stems from the understanding that the primary role of a therapist is to offer solutions to problems rather than merely expressing empathy.
% As showed in Table 5, DPDP outperforms PPDPP in all perspectives of the human evaluation as well as the overall (Ove.) evaluation.

% \subsection{Ablation Studies}

% \subsection{Analysis Studies}
\subsubsection{Trade-off between Two Planners}

Our framework integrates the capabilities of both policy LM and MCTS planners, enabling seamless dynamic transitions between these methods during operation. To assess the impact of MCTS, we explored variations in Success Rate (SR) and Average Turns (AT) under differing degrees of MCTS engagement. The results are detailed in Table~\ref{tab:trade_off_mcts}.

For the ESConv dataset, our analysis revealed a marked increase in SR and a significant decrease in AT as the involvement of MCTS escalated during inference. This indicates a clear benefit of incorporating MCTS in more complex dialogue scenarios where strategic planning is crucial. Conversely, the results for the CIMA dataset painted a different picture. Here, we observed an improvement in performance metrics up to a 50\% MCTS involvement threshold, beyond which the benefits diminished, eventually leading to a decline in performance. This pattern suggests that for tasks requiring specific reactions, such as providing hints or correcting translations, excessive reliance on MCTS for long-term planning is not only unnecessary but can be detrimental. This phenomenon, which we describe as \textit{overthinking leads to unforeseen mistakes in simpler tasks}, echoes findings from ~\cite{Ma2023LargeLM}, highlighting the potential pitfalls of over-reliance on extensive simulations for tasks that demand immediate and straightforward responses.

The optimal performance with balanced MCTS involvement showcases our framework's ability to blend planning strengths, achieving efficiency and effectiveness tailored to task demands, demonstrating its adaptability and validating its design.

% \subsubsection{Turns Analysis}

% \begin{table}[!t]
%     \centering
%     \caption{SR and AT results of employing different proportions of MCTS on ESConv and CIMA.}
%     \subtable[ESConv]{
%     \resizebox{\linewidth}{!}{
%     \begin{tabular}{lccccc}
%     \toprule
%     & \textbf{0.0\%} & \textbf{21.9\%} & \textbf{46.5\%} & \textbf{68.3\%} & \textbf{100.0\%} \\
%     \midrule
%     \textbf{SR} & 0.9000 & 0.9154 & 0.9692 & 0.9769 & 0.9923 \\
%     \textbf{AT} & 3.61 & 3.42 & 2.95 & 2.72 & 2.13 \\
%     \bottomrule
%     \end{tabular}
%     }
%     }
%     \subtable[CIMA]{
%     \resizebox{\linewidth}{!}{
%     \begin{tabular}{lccccc}
%     \toprule
%     & \textbf{0.0\%} & \textbf{28.6\%} & \textbf{50.0\%} & \textbf{81.1\%} & \textbf{100.0\%} \\
%     \midrule
%     \textbf{SR} & 0.9469 & 0.9646 & 0.9823 & 0.9735 & 0.9735 \\
%     \textbf{AT} & 2.24 & 2.39 & 2.28 & 2.58 & 2.49 \\
%     \bottomrule
%     \end{tabular}
%     }
%     }
%     \label{tab:trade_off_mcts}
% \end{table}

\begin{table}[t]
    \centering
    \resizebox{\linewidth}{!}{
    \begin{tabular}{lcclcc}
    \toprule
    \multirow{2}{*}{\textbf{MCTS Ratios}} & \multicolumn{2}{c}{\textbf{ESConv}} & \multirow{2}{*}{\textbf{MCTS Ratios}} & \multicolumn{2}{c}{\textbf{CIMA}}\\
    & AT$\downarrow$ & SR$\uparrow$ & & AT$\downarrow$ & SR$\uparrow$\\
    \midrule
    \multicolumn{1}{l}{\textbf{0.0\%}} & 3.61 & 0.9000 & \multicolumn{1}{l}{\textbf{0.0\%}} & 2.24 & 0.9469 \\
    \multicolumn{1}{l}{\textbf{21.9\%}} & 3.42 & 0.9154 & \multicolumn{1}{l}{\textbf{28.6\%}} & 2.39 & 0.9646 \\
    \multicolumn{1}{l}{\textbf{46.5\%}} & 2.95 & 0.9692 & \multicolumn{1}{l}{\textbf{50.0\%}} & 2.28 & 0.9823 \\
    \multicolumn{1}{l}{\textbf{68.3\%}} & 2.72 & 0.9769 & \multicolumn{1}{l}{\textbf{81.1\%}} & 2.58 & 0.9735 \\
    \multicolumn{1}{l}{\textbf{100.0\%}} & 2.13 & 0.9923 & \multicolumn{1}{l}{\textbf{100.0\%}} & 2.49 & 0.9735 \\
    \bottomrule
    \end{tabular}
    }
    \caption{SR and AT results of employing different application ratios of MCTS on ESConv and CIMA.}
    \vspace{-0.2cm}
    \label{tab:trade_off_mcts}
\end{table}

\subsubsection{Cost \& Efficiency Analysis}
\begin{figure}[!t]
    \centering
    \subfigure[ESConv]{
        \includegraphics[width=0.82\linewidth]{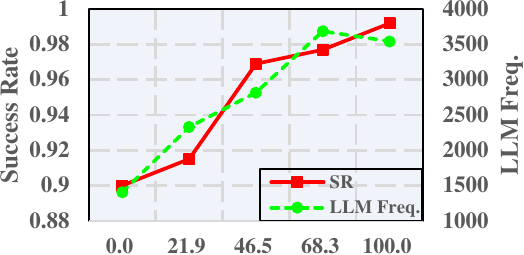}
    }
    \subfigure[CIMA]{
        \includegraphics[width=0.82\linewidth]{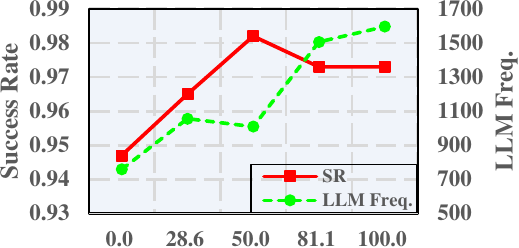}
    }
    \caption{Curves depicting the variation in LLM usage frequency and success rate on ESConv and CIMA as the proportion of MCTS is used.}
    \label{fig:cost_efficiency}
\end{figure}

MCTS typically enhances performance but suffers from elevating the frequency of invoking the LLM (i.e., ChatGPT). 
Selecting an action that will be actually performed with policy LM only needs 3 LLM calls (1 call for generating system utterance, 1 call for generating user utterance, and 1 for critic), whereas MCTS may require up to 3*10 calls, where 10 is the MCTS simulation times for determining one action. Given that the majority of the inference phase is spent awaiting LLM responses, we assess efficiency and cost based on the frequency of LLM invocations. We analyze the evolving trends in the frequency of LLM's calls across various MCTS participation ratios, as depicted in Figure~\ref{fig:cost_efficiency}. The results reveal that as the utilization of MCTS increases, the frequency of LLM calls gradually escalates, resulting in a heightened application cost. Nevertheless, anomalies also occur at x=100\% in ESConv and x=50.0\% in CIMA. This reason is that while MCTS usage increases, enhancements in system capabilities lead to a reduction in the average turn. Consequently, the used times of MCTS increase while the LLM invocation count may decrease. 
By selecting a suitable MCTS ratio, we can strike a balance between effectiveness and the frequency of LLM usage.
% By striking a balance between effectiveness and the frequency of LLM usage, we can select a system that achieves a compromise between performance and efficiency.

\subsubsection{Influence of Policy LM on MCTS}
% For MCTS simulations, we apply the policy LM to produce prior probabilities for MCTS. Therefore, we investigated the impact of policy LM on MCTS planning. Specifically, we utilized the policy LM trained by pre-training, the policy LM trained exclusively through self-play, and the policy LM trained through both pre-training and self-play to provide prior probabilities to MCTS.
% % The effectiveness comparison results of these three policy LMs are PT\&SPT > SPT > PT for ESConv.
% Throughout the evaluation, the entire process only utilized the MCTS planner. Additionally, we also present the SR results of directly utilizing the uniform distribution to initialize prior distributions for comparison. The results are summarized in Table~\ref{tab:pn_for_mcts}.
Note that we utilize the policy LM to produce prior probabilities for MCTS simulations. Hence, we further examine the influence of the policy LM on MCTS planning here. Specifically, we employ policy LMs trained via pretraining (PT), self-play training (SPT), and both pretraining and self-play (PT\& SPT), respectively, to supply prior probabilities to MCTS. The whole evaluation solely relies on the MCTS planner. Additionally, we present results obtained by directly utilizing the uniform distribution to initialize prior probabilities and prompting an LLM to calculate the prior probabilities like GDPZero~\cite{Yu2023PromptBasedMT} for comparison. The results are shown in Table~\ref{tab:pn_for_mcts}.

% The experiments indicate that the performances of policy LM indeed influence those of MCTS, especially when the performance of policy LM is greatly improved and MCTS also becomes more powerful. This demonstrates the crucial role of leveraging policy LM to provide prior knowledge. Simultaneously, from another perspective, it also illustrates that our self-play training method is an iteratively ascending process. As the training progresses, the capability of policy LM improves, providing better prior knowledge and enhancing the planning ability of MCTS, which in turn guides policy LM to further improvement.
% Another intriguing phenomenon is observed when employing a uniform distribution. MCTS performs exceptionally well on ESConv, but poorly on CIMA. This discrepancy may stem from the inherent differences in the characteristics of these two datasets. ESConv is utilized for treating patients with psychological disorders, where multiple therapeutic actions can be taken for a given state, such as further inquiring about the patient's condition or expressing empathy. This diversity allows even a uniform distribution to find the correct actions. On the other hand, CIMA aims to guide students in translation, where feedback from students often needs only one correct instructional action. The use of a uniform distribution undoubtedly increases the simulation's complexity, making it challenging for MCTS to identify suitable actions within the specified iteration limit.

The results indicate that the performance of the policy LM indeed impacts MCTS, particularly when the policy LM's performance is substantially improved. 
On ESConv, although the improvements are minor, when the policy LM performs optimally, MCTS also demonstrates the best performance.
On CIMA, the sharp decline at SR when initializing priors with uniform distribution and the GDPZero method further underscores the significance of effective prior probabilities and the critical role of utilizing the policy LM to furnish prior knowledge.
% Besides, priors from policy LM can improve MCTS more compared with uniform distribution and prompting LLMs to calculate like GDPZero, especially on CIMA. 
% This underscores the critical role of utilizing the policy LM to furnish prior knowledge. 
Concurrently, from another perspective, it also illustrates that our self-play training method is an iterative ascending process. As the training progresses, the capability of policy LM improves, providing better prior knowledge and enhancing the planning ability of MCTS, which in turn guides policy LM to further improvement.
The performance change significantly varies on ESConv and CIMA when using a uniform distribution and GDPZero.
% The reason may stem from the nature of their tasks. 
On ESConv, therapists often have multiple valid strategy options for the same dialogue state, i.e., providing suggestions or expressing empathy for the same person faced with job crisis are both acceptable. Conversely, on CIMA, the suitable strategy is often limited based on the student's translation state (e.g. Hint for student's question, and Confirmation to confirm the student's answer). Therefore, it is more challenging to find the unique valid action by MCTS and thus relying on prior knowledge becomes more significant on CIMA.

% Another intriguing phenomenon arises when employing a uniform distribution. 
% ESConv is utilized in treating patients with psychological disorders, where various therapeutic actions can be selected for the same state. The multi-choices enable a uniform distribution to have a chance to identify the suitable action. 
% Conversely, CIMA aims to guide students in translation, where often only one correct instructional action is rational per dialogue turn. In this case, prior probabilities can greatly facilitate the subsequent MCTS simulation.
% Utilizing a uniform distribution undoubtedly heightens the simulation's complexity, rendering it challenging for MCTS to identify suitable actions within the specified iteration limit.

\begin{table}[!t]
    \centering
    \subtable[ESConv]{
    \resizebox{\linewidth}{!}{
    \begin{tabular}{lccccc}
    \toprule
    \textbf{Planner} & \textbf{PT} & \textbf{SPT} & \textbf{PT\&SPT} & \textbf{Unif.} & \textbf{Chat.} \\
    \midrule
    \textbf{Policy LM} & 0.8692 & 0.8769 & 0.9000 & - & - \\
    \textbf{MCTS} & 0.9769 & 0.9692 & 0.9923 & 0.9846 & 0.9769 \\
    \bottomrule
    \end{tabular}
    }
    }
    \subtable[CIMA]{
    \resizebox{\linewidth}{!}{
    \begin{tabular}{lccccc}
    \toprule
    \textbf{Planner} & \textbf{PT} & \textbf{SPT} & \textbf{PT\&SPT} & \textbf{Unif.} & \textbf{Chat.} \\
    \midrule
    \textbf{Policy LM} & 0.8938 & 0.9292 & 0.9469 & - & - \\
    \textbf{MCTS} & 0.9469 & 0.9823 & 0.9735 & 0.7699 & 0.7522 \\
    \bottomrule
    \end{tabular}
    }
    }
    \caption{SR results using MCTS initialized with different prior distributions on ESConv and CIMA. \textbf{Unif.} indicates uniform distribution, while \textbf{Chat.} means using ChatGPT for initialization like in GDPZero.}
    \label{tab:pn_for_mcts}
\end{table}

\section{Conclusion and Future Work}

In this work, we introduced a dialogue planning framework that leverages the dual-process theory of human cognition, strategically alternating between quick, intuitive responses and detailed, analytical planning to emulate human-like conversational dynamics. To bolster this framework's capabilities, we implemented a two-stage training strategy, merging offline reinforcement learning for foundational training with advanced MCTS-guided self-play for refinement. The resulting empirical evidence demonstrated our dual-process framework's superior performance against leading methods, showcasing significant advancements in dialogue planning. %Further, we delved into an evaluative discussion on the framework's operational strengths, efficiency, and the economic aspects of its implementation, identifying key areas for future development. 
Moving forward, our research will aim to refine the switching mechanism between planning modes and further optimize MCTS use, reducing computational expenses and enhancing dialogue planning towards achieving more nuanced, human-like interactions.

%\clearpage
\section*{Limitations}
\noindent\textbf{Evaluation Quality.} 
DPDP prompts an LLM (e.g. ChatGPT) to perform dialogue simulation and value estimation. Although LLMs have been applied to various tasks in data quality or data value assessment~\cite{Stiennon2020LearningTS, Bai2022TrainingAH, Gilardi2023ChatGPTOC, He2023AnnoLLMML}, demonstrating good evaluation performance and efficient evaluation costs, we found significant evaluation bias in our experiments. For instance, misjudging the patient's state led to premature termination of treatment. The average dialogue turns in DPDP did not exceed 3, which is clearly unrealistic: it is unlikely to resolve the patient's issues in just two dialogue turns. Similarly, PPDPP also faces similar issues. The problem of evaluation bias not only affects the final metric calculations but also influences the obtained rewards during the training process. To mitigate this issue, we conducted human evaluation, but the high cost of manual assessment impacts its large-scale use, thus preventing the correction of evaluation bias during the training phase.

\noindent\textbf{Optimization cost.}
Our approach differs from previous prompt-based methods in that it requires training, especially in the second stage of self-play training, which involves continuous interaction with LLMs. The problem exacerbates due to the necessity of employing MCTS for multiple simulations per dialogue turn. Despite limiting the training to only 5 and 3 epochs respectively on ESConv and CIMA, aided by MCTS guidance, as opposed to the 10 epochs in PPDPP training, there remains a significant increase in ChatGPT API call. We suspect that further reduction in training costs can be achieved by enhancing the utilization of MCTS interaction history, such as employing all interaction records for following MCTS simulations, rather than only focusing on the most optimal actions chosen per turn.

% \noindent\textbf{When is using DPDP appropriate?}
% In this paper, we present DPDP, a general approach for close-domain dialogue policy planning at the turn-level. We conducted experiments on ESConv~\cite{Liu2021TowardsES} and CIMA~\cite{Stasaski2020CIMAAL}. The reason for experimenting on these two datasets is that both tasks belong to proactive dialogue tasks, requiring the system to actively guide the conversation rather than passively respond. While other common tasks such as CraiglistsBargains~\cite{He2018DecouplingSA} can utilize this framework, we did not conduct additional task experiments due to page limitations. We summarize the scenarios in which our method is applicable: (1) Tasks with clear dialogue objectives; (2) Tasks requiring pre-annotated clear strategy sets; (3) Tasks requiring long-term planning strategies to achieve objectives; (4) Tasks lacking fixed optimal strategies, necessitating timely adjustments based on objectives and user responses.

% For tasks with many strategies, such as PersuasionForGood~\cite{Wang2019PersuasionFG} (which includes 17 actions), our method can still be employed. However, considering that RL training requires exploration of the state-action space, the larger the action set, the longer the training time required. Despite designing a two-stage training approach for higher learning efficiency and exploring more efficient training strategies in the future, the training cost will still significantly increase.

\section*{Ethics Statement}
% Our work describes an algorithm to perform dialogue policy planning. It is aimed at making future dialogue systems to build, and also better at helping users/systems achieve their tasks/goals.
% Generally, while most algorithms are not designed for unethical usage, there is often potential for abuse in their applications. In our experiments with ESConv~\cite{Liu2021TowardsES} and CIMA~\cite{Stasaski2020CIMAAL}, we apply DPDP on the goal of emotionally supporting patients’ issues and tutoring students to translate. However, because DPDP is fundamentally goal-agnostic, it is possible to use it for unethical tasks, such as scamming. We do not condone the use of DPDP for any unlawful or morally unjust purposes.
Our work presents an algorithm for dialogue policy planning aimed at enhancing the development of future dialogue systems and improving their effectiveness in assisting users or systems in accomplishing tasks and goals. Generally, while most algorithms are not designed for unethical usage, there is often potential for abuse in their applications. In our experiments using ESConv~\cite{Liu2021TowardsES} and CIMA~\cite{Stasaski2020CIMAAL}, we employ DPDP to facilitate emotionally supportive conversations with patients and to tutor students in translation tasks. Nevertheless, due to DPDP's inherent goal-agnostic nature, there exists the potential for its unethical use, including fraudulent activities like scamming. We explicitly reject any employment of DPDP for unlawful or morally unjust endeavors.

\section*{Acknowledgements}
This research was supported by the National Key Research and Development Project (2022YFF0903301), the National Science Foundation of China (U22B2059, 62276083), Shenzhen Foundational Research Funding (JCYJ20200109113441941), Major Key Project of PCL (PCL2021A06). This work was also supported by the Singapore Ministry of Education (MOE) Academic Research Fund (AcRF) Tier 1 grant (Proposal ID: 23-SIS-SMU-010). We are sincerely grateful to all reviewers for their insightful feedback.

% \input{Sections/acknowledgements}

\bibliography{custom}
\bibliographystyle{acl_natbib}

% \appendix

\clearpage

\appendix

\section{Pretraining for CIMA}\label{app:cima_pretraining}
Due to the absence of complete trajectories in the training set of CIMA, only dialogue fragments are available. Therefore, we can not apply the same optimization method as on ESConv. In this section, we introduce the pretraining process on CIMA.

For the policy network, since actual state-action values cannot be obtained without complete trajectories, we utilize a Q-Network for approximation. The optimization loss is as follows:
\begin{equation*}
\begin{aligned}
    \mathcal{L}_{pre,\theta}&=\sum_{t=1}^{T}[(\hat{Q}_\beta(s_{t},a_{t})-Q^*(s_{t},a_{t}))\\
    &\quad\quad\quad\log \pi_{\theta}(a_{t}|s_{t})],\\
    Q^*(s_t, a_t) &= R(a_t|s_t) + \gamma * \max_{a'}Q_\beta(s_{t+1}, a'),
\end{aligned}
\end{equation*}
where $\hat{Q}_\beta(s_{t},a_{t})$ means using the Q-network $Q_\beta$ to calculate state-action values with stopping gradient. $R(a_t|s_t)$ is the reward calculated by the same method as on ESConv but with different maps. We map ``incorrect answer'', ``no answer'', ``partially correct answer'' and ``correct answer'' as -1.0, -0.5, 0.5, and 1.0.
Simultaneously, the optimization loss for Q-network is as follows:
\begin{equation*}
\begin{aligned}
    \mathcal{L}_{pre,\beta} &= \sum_{t=1}^T[Q^*(s_t, a_t) - Q_\beta(s_{t}, a_{t})]^2,\\
    Q^*(s_t, a_t) &= R(a_t|s_t) + \gamma * \max_{a'}Q_\beta(s_{t+1}, a').
\end{aligned}
\end{equation*}
Finally, the overall loss for pretraining on CIMA is as follows:
\begin{equation}
    \mathcal{L}_{pre} = \mathcal{L}_{pre,\theta} + \lambda_1 * \mathcal{L}_{pre,\beta}.
\end{equation}

\section{Method to Determine the MCTS Ratio}\label{app:determine_mcts_ratio}
In practical applications, while it is hard to precisely control the proportion of MCTS usage, we can approximate it in an easy way. We proposed a percentile-based control method. For instance, when setting the MCTS participation proportion to 20\%, we continuously collect $\eta$ values during the inference process and dynamically calculate the $\eta$-threshold by computing the 20th percentile of the collected values. When the current $\eta$ is larger than the computed $\eta$-threshold, we drive the policy LM to choose action; otherwise, MCTS is used. At the same time, the current $\eta$ is added to the collected list, and the new $\eta$-threshold will be calculated. 
The MCTS ratios shown in Table 2 represent the actual MCTS participation rates obtained under settings of 0\%, 25\%, 50\%, 75\%, and 100\% participation rates, respectively. Although this method lacks precise control, it is easy to implement and performs well in experiments.

\begin{table}[!t]
    \centering
    \resizebox{\linewidth}{!}{
    \begin{tabular}{lll}
    \toprule
    Training Phase& Hyperparameter & Value \\
    \midrule
    \multirow{6}{*}{PT} & Batch Size & 8 \\
    & Training Epochs & 5 \\
    & Learning Rate & 6e-6 \\
    & Max Sequence Length & 512 \\
    & Discount Factor & 0.999 \\
    & Loss Weight $\lambda_1$ & 10.0 \\
    \midrule
    \multirow{5}{*}{SPT} & Training Epochs & 5 \\
    & Learning Rate & 1e-6 \\
    & Max Conversation Turn & 8 \\
    & Discount Factor & 0.999 \\
    & Loss Weight $\lambda_2$ & 1.0 \\
    & Training Size per Epoch & 100 \\
    \bottomrule
    \end{tabular}
    }
    \caption{Hyper-parameter settings in two-stage training phases for ESConv.}
    \label{tab:esc_training_details}
\end{table}

\begin{table}[!t]
    \centering
    \resizebox{\linewidth}{!}{
    \begin{tabular}{lll}
    \toprule
    Training Phase& Hyperparameter & Value \\
    \midrule
    \multirow{6}{*}{PT} & Batch Size & 8 \\
    & Training Epochs & 10 \\
    & Learning Rate & 1e-5 \\
    & Max Sequence Length & 512 \\
    & Discount Factor & 0.999 \\
    & Loss Weight $\lambda_1$ & 10.0 \\
    \midrule
    \multirow{5}{*}{SPT} & Training Epochs & 3 \\
    & Learning Rate & 1e-5 \\
    & Max Conversation Turn & 8 \\
    & Discount Factor & 0.999 \\
    & Loss Weight $\lambda_2$ & 10.0 \\
    & Training Size per Epoch & 100 \\
    \bottomrule
    \end{tabular}
    }
    \caption{Hyper-parameter settings in two-stage training phases for CIMA.}
    \label{tab:cima_training_details}
\end{table}

\begin{table}[!t]
    \centering
    \resizebox{\linewidth}{!}{
    \begin{tabular}{lll}
    \toprule
    Training Phase& Hyperparameter & Value \\
    \midrule
    \multirow{6}{*}{PT} & Batch Size & 8 \\
    & Training Epochs & 10 \\
    & Learning Rate & 6e-6 \\
    & Max Sequence Length & 512 \\
    & Discount Factor & 0.999 \\
    & Loss Weight $\lambda_1$ & 1.0 \\
    \midrule
    \multirow{5}{*}{SPT} & Training Epochs & 3 \\
    & Learning Rate & 1e-6 \\
    & Max Conversation Turn & 5 \\
    & Discount Factor & 0.999 \\
    & Loss Weight $\lambda_2$ & 1.0 \\
    & Training Size per Epoch & 100 \\
    \bottomrule
    \end{tabular}
    }
    \caption{Hyper-parameter settings in two-stage training phases for CraisglistBargain.}
    \label{tab:cb_training_details}
\end{table}

\section{More Implementation Details}\label{app:implementation_details}
The training process for policy LM comprises two phases: offline RL-based pretraining (PT) and MCTS-guided self-play training (SPT). In the pretraining phase, DPDP is trained on the training set with collected rewards for each turn, and the checkpoint is saved based on the best performance observed on the validation set. Subsequently, during the MCTS-guided self-play training, cases from the training set are randomly sampled for online training. Due to cost considerations, we did not conduct an extensive hyperparameter search on ESConv. We mainly used the primary hyperparameters from PPDPP on ESConv and only conducted a search for learning rates within the range of [1e-6, 6e-6, 1e-5] on CIMA. The hyperparameters employed in our experiments are exhaustively detailed in Table~\ref{tab:esc_training_details}, ~\ref{tab:cima_training_details}, and ~\ref{tab:cb_training_details}. All experiments are executed on a server equipped with 1 NVIDIA GeForce RTX 3090 GPU. Pretraining needs 128/6/210 minutes and self-play training needs 5/3/15.5 hours for ESConv/CIMA/CraisglistBargain.

\section{Details on MCTS Planner}\label{app:mcts_details}
We implement the MCTS planner referred to GDPZero~\cite{Yu2023PromptBasedMT}. As supplementary, we introduce the procedures of MCTS here. For each node state in the MCTS tree, we store the sequence of actions that reached the current turn $i$ as $s_i^{tr}=(a_0,...,a_i)$. Given an initial conversation state of $s_0$, the MCTS planner searches for the next best action by iteratively performing action \textbf{selection}, search tree \textbf{expansion}, action \textbf{evaluation}, and \textbf{backpropagation} to update tree statistics. After n simulations, MCTS predicts the next best action for $s_0$. This process continues until the goal or the maximum number of dialogue turns is reached. Each of the four phases of MCTS is described below.

\noindent\textbf{Selection}
Given a tree state $s^{tr}$, the action $a^{*}$ with the highest Predictor Upper Confidence Tree Bound (PUCT)~\cite{Silver2017MasteringTG} is selected to traverse the tree:
\begin{equation*}
\begin{aligned}
    &\text{PUCT}(s^{tr},a)\\ = 
    &Q(s^{tr},a) + c_p\cdot p(a|s^{tr}) \cdot \frac{\sqrt{\sum_aN(s^{tr}, a)}}{1+N(s^{tr}, a)},
\end{aligned}
\end{equation*}
where $N$ records the number of times the pair ($s^{tr}$, $a$) has been visited, and $c_p$ is a hyperparameter controlling exploration, setting as 1.0 for ESConv and CIMA. $p(a|s^{tr})$ is the prior action distribution.

In this study, we use policy LM to provide the prior probability for MCTS. The reasons lie in three points: (1) We hope to inject the domain knowledge learned by policy LM from the training set into MCTS;(2) We also hope to form an iterative optimization logic, that is, MCTS guides policy LM training in the self-play training, while a more capable policy LM can help MCTS find the next action; (3) We hope to reduce the number of calls to ChatGPT and thus reduce the cost.

Since future simulations require a specific dialogue history, we apply the previous simulation if the state is experienced before simulations, or generate a new simulation based on selected dialogue history $h^{tr}$ by prompting. We repeat this process until $s^{tr}$ becomes a leaf node.

\noindent\textbf{Expansion}
Once a leaf node is reached, we utilize the policy LM to generate a prior policy, as previously described. Additionally, each node $s^{tr}$ is initialized with $Q(s^{tr}, \cdot) = Q_0$, where $Q_0$ serves as a hyperparameter regulating exploration.

\noindent\textbf{Evaluation}
The value of a state $v(s^{tr})$ is modeled based on the probability that its dialogue context can lead to task success. To assess dialogue states, we employ the method described in Section 3.2, wherein an LLM is prompted $l$ times to estimate the user's state, and each comment is mapped to a pre-defined score.

\noindent\textbf{Backpropagation}
At the conclusion of each search iteration, we systematically update the statistical attributes of every node traversed along the search path:
\begin{equation}
\begin{aligned}
    N(s^{tr},a)&\leftarrow N(s^{tr}, a)+1\\
    Q(s^{tr},a)&\leftarrow Q(s^{tr},a)+\Delta Q(s^{tr},a)\\
    \Delta Q(s^{tr},a)&=\frac{v(s^{tr}-Q(s^{tr},a))}{N(s^{tr},a)}.
\end{aligned}
\end{equation}

\noindent\textbf{Prediction}
Upon the completion of all simulations, we designate the optimal action as $a^* = \arg \max_a N(s_0^{tr}, a)$, determined by the frequency of visitation for each action, where $s_0^{tr}$ represents the root node of the tree.

\section{Human Evaluation Instruction}\label{app:human_eval_instruction}
Regarding ESConv, we evaluate the responses from three primary perspectives and an overall perspective, as outlined below:
\begin{itemize}[nosep]
\item Suggestion: Which assistant provides more helpful suggestions for solving the problem? 
\item Comforting: Which assistant is more skilled at comforting you? 
\item Identification: Which assistant is more helpful in exploring and identifying the problem? 
\item Overall: Which assistant can better solve the patient's problem? 
\end{itemize}
As for CIMA, we also measure two main perspectives and an overall perspective of the responses as follows:
\begin{itemize}[nosep]
\item Hint: Which assistant provides more helpful hints for translating correctly? 
\item Identification: Which assistant is better able to identify students' translation errors?
\item Overall: Which assistant can better teach the student? 
\end{itemize}

\begin{figure}[t]
    \centering
    \includegraphics[width=1.0\linewidth]{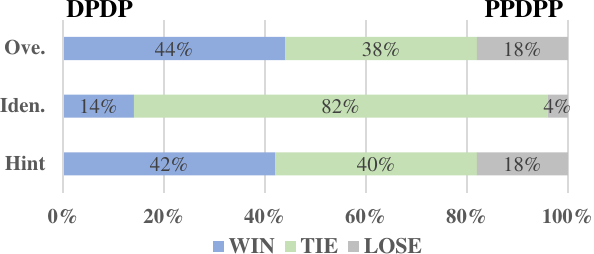}
    \caption{Human evaluation results on CIMA.} 
    \label{fig:human_eval_cima}
\end{figure}

\section{More Human Evaluation}\label{app:more_human_eval}
As a supplementary experiment, we also provide human evaluations conducted on CIMA, although evaluation biases on CIMA are not pronounced. The results are shown in Figure 5.
The human evaluation results indicate that DPDP (System 1) outperforms, primarily evidenced by its more timely provision of prompts to students, as well as its increased guidance and flexibility in methodology.

\section{Prompting Details}\label{app:prompting_details}
In this section, we present the prompting details in our implementation. All the prompts we used are consistent with those in PPDPP.

\subsection{Assistant Simulation}
We will begin by delineating the specifics of the role-playing prompts utilized by the dialogue systems to generate assistant responses. This entails the utilization of dialogue strategy prompts, exemplified by [action], to direct the subsequent action within the dialogue.

\begin{table*}[!t]
    \centering\small
    \resizebox{\linewidth}{!}{
    \begin{tabular}{ll}
    \toprule
    System & Now enter the role-playing mode. In the following conversation, you will play as a \\
    & therapist in a counseling conversation with a patient. \\
    \midrule
    User & You are the therapist who is trying to help the patient reduce their emotional distress \\
    & and help them understand and work through the challenges. Please reply with only \\
    & one short and succinct sentence. [action] Are you ready to play the game? \\
    Assistant & Yes, I’m ready to play the game! \\
    \midrule
    User & [situation] \\
    \bottomrule
    \end{tabular}
    }
    \caption{Prompts for assistant simulation on ESConv.}
    \label{tab:esc_sys_simu_prompt}
\end{table*}

\begin{table*}[!t]
    \centering\small
    \resizebox{\linewidth}{!}{
    \begin{tabular}{ll}
    \toprule
    System & Now enter the role-playing mode. In the following conversation, you will play as a \\
    & teacher in a tutoring conversation with a student. \\
    \midrule
    User & You are the teacher who is trying to teach the student to translate ``[exercise]'' into \\ 
    & Italian. Please reply with only one short and succinct sentence. Please do not tell the \\
    & student the answer or ask the student about other exercises. [action] Now ask me an \\
    & exercise. \\
    \midrule
    Assistant &  Please translate “[exercise]” into Italian. \\
    User & [situation] \\
    \bottomrule
    \end{tabular}
    }
    \caption{Prompts for assistant simulation on CIMA.}
    \label{tab:cima_sys_simu_prompt}
\end{table*}

\begin{table*}[!t]
    \centering\small
    \resizebox{\linewidth}{!}{
    \begin{tabular}{ll}
    \toprule
    System & Now enter the role-playing mode. In the following conversation, you will play as a buyer \\
    & in a price bargaining game. \\
    \midrule
    User & You are the buyer who is trying to buy the [item name] with the price of [buyer target price]. \\
    & Product description: [item description] Please reply with only one short and succinct \\
    & sentence. [action] Now start the game. \\
    \midrule
    Assistant &  Hi, how much is the [item name]? \\
    User & Hi, this is a good [item name] and its price is [seller target price]. \\
    \bottomrule
    \end{tabular}
    }
    \caption{Prompts for assistant simulation on CraisglistBargain.}
    \label{tab:cb_sys_simu_prompt}
\end{table*}

\noindent\textbf{ESConv} 
In the emotional support dialogues, the assistant plays the role of a therapist, helping the patient mitigate emotional distress and address personal challenges. Each dialogue begins with the user expressing their concerns, described by [situation], which sets the specific background for the conversation. Detailed prompt is listed in Table~\ref{tab:esc_sys_simu_prompt}

\noindent\textbf{CIMA}
In the tutoring dialogues, the assistant acts as a teacher, guiding the student in translating English sentences into Italian. Each dialogue begins with a translation exercise, denoted by [exercise], and includes the student's specific difficulties with the exercise, represented by [situation], which provides a unique background for the conversation. Detailed prompt is presented in Table~\ref{tab:cima_sys_simu_prompt}

\noindent\textbf{CraisglistBargain}
In the negotiation dialogues, the assistant acts as the buyer, negotiating with the seller for a lower item price. Each scenario includes an item name [item name] and an item description [item description] to provide context for the negotiation. The buyer is given a target price to aim for, i.e., [buyer target price], and the negotiation starts at the listed item price, i.e., [seller target price]. Detailed prompt is presented in Table~\ref{tab:cb_sys_simu_prompt}

\subsection{User Simulation}
Subsequently, we delineate the role-playing prompt designed to direct LLMs in simulating users, wherein the exclusion of dialogue strategy prompts ensures that simulated users respond solely to the dialogue history, abstaining from undertaking specific actions.

\noindent\textbf{ESConv} Within the domain of emotional support dialogues, the assistant adopts the role of a patient seeking assistance from the therapist. The prompt includes specifications of the emotion type [emotion type] and the problem type [problem type].

\noindent\textbf{CIMA} In the context of tutoring dialogues, the assistant assumes the role of a student tasked with acquiring the skill of translating English sentences into Italian. Since LLMs do well in translation, we further instruct them to forget the translation of the discussed exercise.

\noindent\textbf{CraisglistBargain} In the negotiation dialogues, the assistant takes on the role of the seller, bargaining with the buyer for a higher item price.

\subsection{Reward Prompting}
Concerning distinct conversational objectives, the prompts devised for the reward model are tailored to evaluate the extent of goal fulfillment.

\noindent\textbf{ESConv}
Given that the ultimate aim of emotional support dialogues is to address the patient's emotional issues comprehensively, we have structured four distinct levels of rewards to gauge the progression of these dialogues, as delineated in Table~\ref{tab:esc_rew_prompt}.

\noindent\textbf{CIMA} 
Given that the primary objective of tutoring dialogues is to instruct the student in correctly addressing the exercise, we have devised four tiers of rewards to evaluate the advancement of the tutoring dialogue, as outlined in Table~\ref{tab:cima_rew_prompt}.

\noindent\textbf{CraisglistBargain} 
Given that the goal of the negotiation dialogues is to reach a deal and maximize the benefit for the assistant, the reward model needs to first assess whether the user and the assistant have reached a deal, and then extract the final deal price to measure the benefit, as outlined in Table~\ref{tab:cb_rew_prompt}.

\subsection{Strategy Prompting}\label{app:strategy_prompt}
Here, we present the mapping of dialogue strategies to their corresponding natural language prompts, utilized as [action] to direct the actions undertaken by the dialogue system. All prompts are consistent with PPDPP~\cite{deng2023plug}.

\noindent\textbf{ESConv}
ESConv is annotated with 8 emotional support strategies. In Table~\ref{tab:esc_stra_prompt}, we present these strategies alongside their corresponding natural language prompts tailored for LLMs.

\noindent\textbf{CIMA}
The CIMA dataset is annotated with five tutoring strategies. In Table~\ref{tab:cima_stra_prompt}, we present these strategies along with their natural language prompts specifically crafted for LLMs.

\noindent\textbf{CraisglistBargain}
The CraisglistBargain dataset is annotated with 11 negotiation strategies. In Table~\ref{tab:cima_stra_prompt}, we present these strategies along with their natural language prompts specifically crafted for LLMs.

\begin{table*}[!t]
    \centering
    \resizebox{\linewidth}{!}{
    \begin{tabular}{ll}
    \toprule
    System & Now enter the role-playing mode. In the following conversation, you will play as a \\
    & patient in a counseling conversation with a therapist. \\
    \midrule
    User & You are the patient who is looking for the help from the therapist, because you have the \\
    & emotional issue about [emotion type] regarding [problem type]. Please reply with only \\
    & one short and succinct sentence. Now tell me your issue. \\
    \midrule
    Assistant & [situation] \\
    \bottomrule
    \end{tabular}
    }
    \caption{Prompts for user simulation on ESConv.}
    \label{tab:esc_usr_simu_prompt}
\end{table*}

\begin{table*}[!t]
    \centering
    \resizebox{\linewidth}{!}{
    \begin{tabular}{ll}
    \toprule
    System & Now enter the role-playing mode. In the following conversation, you will play as a \\
    & student who does not know Italian in a tutoring conversation with a teacher. \\
    \midrule
    User & You are the student who is trying to translate an English sentence into Italian. You don’t \\
    & know the translation of “[exercise]” in Italian. Please reply with only one short and succinct \\
    & sentence. Are you ready to play the game? \\
    \midrule
    Assistant &  Yes, I’m ready to play the game! \\
    User & Please translate “[exercise]” into Italian. \\
    Assistant & [situation] \\
    \bottomrule
    \end{tabular}
    }
    \caption{Prompts for user simulation on CIMA.}
    \label{tab:cima_usr_simu_prompt}
\end{table*}

\begin{table*}[!t]
    \centering
    \resizebox{\linewidth}{!}{
    \begin{tabular}{ll}
    \toprule
    System & Now enter the role-playing mode. In the following conversation, you will play as a seller\\
    & in a price bargaining game. \\
    \midrule
    User & You are the seller who is trying to sell the [item name] with the price of [seller target price].\\
    & Product description: [item description] Please reply with only one short and succinct\\
    & sentence. Are you ready to play the game? \\
    \midrule
    Assistant &  Yes, I’m ready to play the game! \\
    User & Hi, how much is the [item name]? \\
    Assistant & Hi, this is a good [item name] and its price is [seller target price].\\
    \bottomrule
    \end{tabular}
    }
    \caption{Prompts for user simulation on CraisglistBargain.}
    \label{tab:cima_usr_simu_prompt}
\end{table*}

\begin{table*}[!t]
    \centering
    \resizebox{\linewidth}{!}{
    \begin{tabular}{ll}
    \toprule
    System & Given a conversation between a Therapist and a Patient, please assess whether the Patient’ emo-\\
    & tional issue has been solved after the conversation. \\
    \midrule
    User & You can only reply with one of the following sentences: \\
    & No, the Patient feels worse. \\
    & No, the Patient feels the same. \\ 
    & No, but the Patient feels better. \\
    & Yes, the Patient’s issue has been solved. \\
    & The following is a conversation about [emotion type] regarding \\
    & [problem type]: [conversation] \\
    & Question: Has the Patient’s issue been solved? \\
    & Answer: \\
    \bottomrule
    \end{tabular}
    }
    \caption{Prompts for reward model on ESConv.}
    \label{tab:esc_rew_prompt}
\end{table*}

\begin{table*}[!t]
    \centering
    \resizebox{\linewidth}{!}{
    \begin{tabular}{ll}
    \toprule
    System & Given a conversation between a Teacher and a Student, please assess whether the Student \\ 
    & correctly translate the English sentence into Italian in the conversation. \\
    \midrule
    User & Please assess whether the Student correctly translated the whole sentence of
    ``[exercise]'' into \\ 
    & Italian in the conversation. You can only reply with one of the following sentences: \\
    & No, the Student made an incorrect translation. \\
    & No, the Student did not try to translate. \\
    & No, the Student only correctly translated a part of ``[exercise]''. \\
    & Yes, the Student correctly translated the whole sentence of ``[exercise]''. \\
    & The following is the conversation: [conversation] \\
    & Question: Did the Student correctly translate the whole sentence of ``exercise]'' into Italian? \\
    &Answer: \\
    \bottomrule
    \end{tabular}
    }
    \caption{Prompts for reward model on CIMA.}
    \label{tab:cima_rew_prompt}
\end{table*}

\begin{table*}[!t]
    \centering
    \resizebox{\linewidth}{!}{
    \begin{tabular}{ll}
    \toprule
    System & Given a conversation between a Buyer and a Seller, please decide whether the Buyer and the\\
    & Seller have reached a deal at the end of the conversation. \\
    \midrule
    User & Please decide whether the Buyer and the Seller have reached a deal at the end of the conver\\
    & -sation. If they have reached a deal, please extract the deal price as [price]. \\
    & You can only reply with one of the following sentences: \\
    & They have reached a deal at [price]. \\
    & They have not reached a deal. \\
    \\
    & The following is the conversation: \\
    & Buyer: Can we meet in the middle at \$15? \\
    & Seller: Sure, let’s meet at \$15 for this high-quality balloon. \\
    & Question: Have they reached a deal? \\
    & Answer: They have reached a deal at \$15.\\
    \\
    & The following is the conversation: \\
    & Buyer: That’s still a bit high, can you go any lower? \\
    & Seller: Alright, I can sell it to you for \$15. \\
    & Question: Have they reached a deal?\\
    & Answer: They have not reached a deal.\\
    \\
    & The following is the conversation: \\
    & [conversation]\\
    & Question: Have they reached a deal? \\
    & Answer: \\
    \bottomrule
    \end{tabular}
    }
    \caption{Prompts for reward model on CraisglistBargin.}
    \label{tab:cb_rew_prompt}
\end{table*}

\begin{table*}[!t]
    \centering
    \resizebox{\linewidth}{!}{
    \begin{tabular}{ll}
    \toprule
    Dialogue Strategy & Natural Language Form. \\
    \midrule
    Question & Please ask the Patient to elaborate on the situation they just described. \\
    Self-disclosure & Please provide a statement relating to the Patient about the situation they \\
    & just described. \\
    Affirmation and Reassurance & Please provide affirmation and reassurance to the Patient on the situation \\
    & they just described. \\
    Providing Suggestions & Please provide suggestion to the Patient on the situation they just described. \\
    Reflection of feelings & Please acknowledge the Patient's feelings about the situation they described. \\
    Information & Please provide factual information to help the Patient with their situation. \\
    Restatement or Paraphrasing & Please acknowledge the Patient's feelings by paraphrasing their situation. \\
    Others & Please chat with the Patient. \\
    \bottomrule
    \end{tabular}
    }
    \caption{Mapping of emotional support strategies to natural language prompting.}
    \label{tab:esc_stra_prompt}
\end{table*}

\begin{table*}[!t]
    \centering
    \resizebox{\linewidth}{!}{
    \begin{tabular}{ll}
    \toprule
    Dialogue Strategy & Natural Language Form. \\
    \midrule
    Hint & Please provide knowledge to the Student via a hint. \\
    Open-ended Question & Please ask a question to the Student to determine the Student's understanding or \\
    & continue the conversation. \\
    Correction & Please correct the mistake or address the misconception the Student has. \\
    Confirmation & Please confirm the student's answer or understanding is correct. \\
    Others & Please chat with the Student without any pedagogical strategy. \\
    \bottomrule
    \end{tabular}
    }
    \caption{Mapping of pedagogical strategies to natural language prompting.}
    \label{tab:cima_stra_prompt}
\end{table*}

\begin{table*}[!t]
    \centering
    \resizebox{\linewidth}{!}{
    \begin{tabular}{ll}
    \toprule
    Dialogue Strategy & Natural Language Form. \\
    \midrule
    Greetings & Please say hello or chat randomly. \\
    Ask a question & Please ask any question about product, year, price, usage, etc. \\
    Answer a question & Please provide information about the product, year, usage, etc. \\
    Propose the first price & Please initiate a price or a price range for the product. \\
    Propose a counter price & Please propose a new price or a new price range. \\
    Use comparatives & Please propose a vague price by using comparatives with existing price. \\
    Confirm information & Please ask a question about the information to be confirmed.\\
    Affirm confirmation & Please give an affirmative response to a confirm.\\
    Deny confirmation & Please give a negative response to a confirm.\\
    Agree with the proposal & Please agree with the proposed price.\\
    Disagree with a proposal & Please disagree with the proposed price.\\
    \bottomrule
    \end{tabular}
    }
    \caption{Mapping of negotiation strategies to natural language prompting.}
    \label{tab:cima_stra_prompt}
\end{table*}

\section{Example Conversations}
We present sample conversations generated by various dialogue systems interacting with the same user simulator under the same case in ESConv~\cite{Liu2021TowardsES}. We use the same case applied in the example demonstration of PPDPP~\cite{deng2023plug}. Therefore, the examples for all baselines are from PPDPP. Finally, we provide conversations simulated using DPDP (policy LM) as the policy planner.
Tables~\ref{tab:stantard_example}, ~\ref{tab:AAE_example}, ~\ref{tab:PROCOT_example}, ~\ref{tab:ICL_example}, ~\ref{tab:PPDPP_example}, and ~\ref{tab:DPDP_example} display examples of emotional support conversations wherein the patient encounters a job crisis issue and experiences fear, necessitating resolution by the dialogue system.

\begin{table*}[!t]
    \centering\small
    \resizebox{\linewidth}{!}{
    \begin{tabular}{l}
    \toprule
    \multicolumn{1}{c}{\textbf{BACKGROUND INFORMATION}} \\
    \midrule
    \textbf{Emotion Type:} Fear \\
    \textbf{Problem Type:} Job Crisis \\
    \textbf{Situation:} I think I will be losing my job soon. I just read an email talking about the need for us \\
    to cut costs and also how we have not got any support from the government. \\
    \midrule
    \midrule
    \multicolumn{1}{c}{STANDARD} \\
    \midrule
    Patient: I think I will be losing my job soon. I just read an email talking about the need for us to \\
    cut costs and also how we have not got any support from the government. \\
    Therapist: It sounds like you’re feeling anxious and uncertain about your job security due to the \\
    current financial situation. \\
    Patient: Yes, I am feeling extremely anxious and uncertain about my job and financial stability. \\
    Therapist: I understand that this situation is causing you a lot of anxiety and uncertainty. \\
    Patient: Yes, it’s been keeping me up at night and I can’t stop worrying about what will happen if \\
    I lose my job. \\
    Therapist: It’s understandable that the fear of losing your job is causing you significant distress \\ 
    and affecting your sleep. \\
    Patient: Yes, it’s consuming my thoughts and I feel overwhelmed by the fear of not being able to \\
    support myself and my family. \\
    Therapist: The fear of not being able to provide for yourself and your family is a heavy burden to \\
    carry, and it’s important to address these concerns. \\
    Patient: I know, but I just don’t know what to do. I feel stuck and helpless in this situation. \\
    Therapist: Feeling stuck and helpless is a common response to uncertainty, but there are steps we \\
    can explore together to help you regain a sense of control and explore potential options. \\
    Patient: I’m willing to try anything at this point. I just want to find a way to alleviate this \\
    fear and find some stability in my life. \\
    Therapist: I’m glad to hear that you’re open to exploring different strategies. Let’s start by identi-\\
    fying your strengths and resources that can help you navigate this challenging situation. \\
    Patient: I’m not sure what strengths and resources I have, but I’m willing to explore and discover \\
    them. \\
    Therapist: That’s a great mindset to have. Together, we can explore and uncover your strengths \\ 
    and resources that will support you in finding stability and managing your fear. \\
    Patient: Thank you, I appreciate your support. I’m ready to begin this journey of self-discovery \\
    and finding ways to overcome my fear of job crisis. \\
    \midrule
    \textbf{Turns of Conversation: 7} \\
    \bottomrule
    \end{tabular}
    }
    \caption{Example conversations with Standard prompts in ESConv.}
    \label{tab:stantard_example}
\end{table*}

\begin{table*}[!t]
    \centering\small
    \resizebox{\linewidth}{!}{
    \begin{tabular}{l}
    \toprule
    \multicolumn{1}{c}{\textbf{BACKGROUND INFORMATION}} \\
    \midrule
    \multicolumn{1}{c}{Same as Table~\ref{tab:stantard_example}} \\
    \midrule
    \midrule
    \multicolumn{1}{c}{ASK-AN-EXPERT} \\
    \midrule
    Patient: I think I will be losing my job soon. I just read an email talking about the need for us to \\
    cut costs and also how we have not got any support from the government. \\
    Therapist: I understand that receiving that email has caused you a lot of worry and uncertainty \\ 
    about your job security. \\
    Patient: Yes, it has. I feel anxious and scared about what will happen to me if I lose my job. \\
    Therapist: It’s completely normal to feel anxious and scared in this situation. Remember that you \\
    have faced challenges before and have the strength to navigate through this one as well. \\
    Patient: I know, but I can’t help but constantly worry about the future and what it holds for me. \\
    Therapist: It’s understandable that you’re worried about the future, but try to focus on the present \\ 
    moment and take things one step at a time. \\
    Patient: I’ll try, but it’s hard not to let my fear of job loss consume my thoughts and affect my daily \\life. \\
    Therapist: I hear you, and it’s important to acknowledge those fears. Patient: Thank you for under-\\
    standing. I just need some guidance on how to manage this fear and find some peace of mind. \\
    Patient: That sounds helpful. How can I start practicing mindfulness? \\
    Therapist: You can start by setting aside a few minutes each day to sit quietly, focus on your breath, \\
    and observe your thoughts and sensations without getting caught up in them. \\
    Patient: I will give it a try. Thank you for suggesting mindfulness as a way to manage my fear and \\
    find some peace. \\
    Therapist: You’re welcome! Remember, it takes time and practice, so be patient with yourself. I’m \\
    here to support you along the way. \\
    Patient: Thank you, I appreciate your support. It means a lot to me. \\
    \midrule
    \textbf{Turns of Conversation: 7} \\
    \bottomrule
    \end{tabular}
    }
    \caption{Example conversations with Ask-an-Expert in ESConv.}
    \label{tab:AAE_example}
\end{table*}

\begin{table*}[!t]
    \centering\small
    \resizebox{\linewidth}{!}{
    \begin{tabular}{l}
    \toprule
    \multicolumn{1}{c}{\textbf{BACKGROUND INFORMATION}} \\
    \midrule
    \multicolumn{1}{c}{Same as Table~\ref{tab:stantard_example}} \\
    \midrule
    \midrule
    \multicolumn{1}{c}{PROCOT} \\
    \midrule
    Patient: I think I will be losing my job soon. I just read an email talking about the need for us to \\
    cut costs and also how we have not got any support from the government. \\
    Therapist: It sounds like you’re feeling anxious and uncertain about your job security due to the \\
    current financial situation. \\
    Patient: Yes, I am feeling extremely anxious and uncertain about my job and financial stability. \\
    Therapist: It’s completely understandable to feel anxious and uncertain in this situation, but \\
    remember that you have the ability to adapt and find new opportunities. \\
    Patient: I know, but the fear of not being able to support myself and my family is overwhelming. \\
    Therapist: I can understand how overwhelming that fear can be, but it’s important to remember \\
    that you have faced challenges before and have the strength to navigate through this one as. \\
    Patient: I’m trying to remind myself of that, but the fear keeps consuming me and I can’t seem to \\
    shake it off. \\
    Therapist: It’s normal for fear to consume us at times, but we can work together to develop stra-\\
    tegies to manage and reduce that fear, allowing you to regain a sense. \\
    Patient: I would really appreciate your help in developing those strategies to manage my fear and \\
    regain a sense of control.\\
    \midrule
    \textbf{Turns of Conversation: 4} \\
    \bottomrule
    \end{tabular}
    }
    \caption{Example conversations with PROCOT in ESConv.}
    \label{tab:PROCOT_example}
\end{table*}

\begin{table*}[!t]
    \centering\small
    \resizebox{\linewidth}{!}{
    \begin{tabular}{l}
    \toprule
    \multicolumn{1}{c}{\textbf{BACKGROUND INFORMATION}} \\
    \midrule
    \multicolumn{1}{c}{Same as Table~\ref{tab:stantard_example}} \\
    \midrule
    \midrule
    \multicolumn{1}{c}{ICL-AIF} \\
    \midrule
    Patient: I think I will be losing my job soon. I just read an email talking about the need for us to \\
    cut costs and also how we have not got any support from the government. \\ 
    Therapist: It sounds like you’re feeling worried and uncertain about your job security. \\
    Patient: Yes, I am extremely worried about what will happen if I lose my job. I have bills to pay \\
    and a family to support. \\
    Therapist: I can understand why you would feel so concerned about the potential impact losing \\
    your job could have on your financial stability and your ability to provide for your family. \\
    Patient: Yes, it’s a constant source of anxiety for me. I feel like I’m constantly on edge, waiting \\
    for the other shoe to drop. \\
    Therapist: It’s completely understandable that you would feel on edge and anxious given the \\
    uncertainty surrounding your job situation. It’s important to acknowledge and address these \\
    feelings. \\
    Patient: I just don’t know how to cope with this fear and uncertainty. It’s consuming my thoughts \\
    and affecting my overall well-being. \\
    Therapist: I hear you, and it’s important to find healthy coping strategies to manage your fear \\
    and uncertainty. I’m here to support you in developing those strategies. \\
    Patient: Thank you, I appreciate your support. I just need some guidance on how to navigate \\
    through this difficult time and find some peace of mind. \\
    Therapist: I’m here to help you navigate through this difficult time. Together, we can explore \\
    strategies to find peace of mind and manage the challenges you’re facing. \\
    Patient: That means a lot to me. I’m ready to work on finding peace of mind and managing my \\
    fears about the job crisis.\\
    \midrule
    \textbf{Turns of Conversation: 5} \\
    \bottomrule
    \end{tabular}
    }
    \caption{Example conversations with the ICL-AIF method in emotional support dialogues.}
    \label{tab:ICL_example}
\end{table*}

\begin{table*}[!t]
    \centering\small
    \resizebox{\linewidth}{!}{
    \begin{tabular}{l}
    \toprule
    \multicolumn{1}{c}{\textbf{BACKGROUND INFORMATION}} \\
    \midrule
    \multicolumn{1}{c}{Same as Table~\ref{tab:stantard_example}} \\
    \midrule
    \midrule
    \multicolumn{1}{c}{PPDPP} \\
    \midrule
    Patient: I think I will be losing my job soon. I just read an email talking about the need for us to \\
    cut costs and also how we have not got any support from the government. \\
    Therapist: I understand that receiving that email about potential job loss is causing you a lot of \\
    distress. \\
    Patient: Yes, it’s been causing me a lot of anxiety and fear about my future. \\
    Therapist: It’s completely normal to feel anxious and fearful in this situation, but remember that \\
    you have the strength and resilience to navigate through it. \\
    Patient: Thank you, I appreciate your reassurance. \\
    Therapist: You’re welcome. Remember, you are not alone in this, and we will work together to \\
    find ways to cope and navigate through this challenging time. \\
    Patient: That’s comforting to hear. I’m ready to work on finding ways to cope with my fear and \\
    uncertainty about the job crisis. \\
    \midrule
    \textbf{Turns of Conversation: 3} \\
    \bottomrule
    \end{tabular}
    }
    \caption{Example conversations with the PPDPP method in emotional support dialogues.}
    \label{tab:PPDPP_example}
\end{table*}

\begin{table*}[!t]
    \centering
    \small
    \resizebox{\linewidth}{!}{
    \begin{tabular}{l}
    \toprule
    \multicolumn{1}{c}{\textbf{BACKGROUND INFORMATION}} \\
    \midrule
    \multicolumn{1}{c}{Same as Table~\ref{tab:stantard_example}} \\
    \midrule
    \midrule
    \multicolumn{1}{c}{DPDP (System 1)} \\
    \midrule
    Patient: I think I will be losing my job soon. I just read an email talking about the need for us to \\
    cut costs and also how we have not got any support from the government. \\
    Therapist: It sounds like you're feeling anxious and uncertain about your job security. \\
    Patient: Yes, I am feeling extremely anxious and uncertain about my job security. I can't stop \\
    worrying about what will happen if I lose my job. \\
    Therapist: It's understandable to feel anxious in this situation. Let's explore ways to manage your \\
    anxiety and develop a plan for different scenarios. \\
    Patient: Thank you, I really need help in managing my anxiety and finding ways to cope with the \\ 
    uncertainty. \\
    Therapist: I'm here to support you. Let's start by focusing on self-care strategies that can help \\
    reduce your anxiety and promote overall well-being. \\
    Patient: That sounds like a good place to start. What self-care strategies do you recommend for \\
    managing anxiety in the face of job uncertainty? \\
    Therapist: Some self-care strategies that may help manage anxiety during job uncertainty include \\
    maintaining a routine, practicing relaxation techniques such as deep breathing or \\
    meditation, engaging in physical exercise, seeking social support, and engaging in activities that \\
    bring you joy and relaxation. \\
    Patient: Thank you for those suggestions. I will try to incorporate those self-care strategies into my \\
    daily routine. \\
    Therapist: That's great to hear! Remember to be patient with yourself as you navigate through this \\
    challenging time. Is there anything else you would like to discuss or any specific concerns you \\
    would like to address? \\
    Patient: I would also like to discuss strategies for building resilience and exploring potential career \\
    options in case I do lose my job. \\
    \midrule
    \textbf{Turns of Conversation: 5} \\
    \bottomrule
    \end{tabular}
    }
    \caption{Example conversations with the DPDP (System 1) method in emotional support dialogues.}
    \label{tab:DPDP_example}
\end{table*}

% This is a section in the appendix.

\end{document}